\documentclass[nohyperref]{article}

\usepackage{microtype}
\usepackage{graphicx}
\usepackage{subfigure}
\usepackage{booktabs} 

\usepackage{hyperref}

\usepackage[accepted]{icml2022}

\usepackage{amsmath}\allowdisplaybreaks
\usepackage{amsthm}
\usepackage{amssymb}
\usepackage{graphicx}
\usepackage{xcolor}
\usepackage{bm}
\usepackage[capitalize,noabbrev]{cleveref}
\usepackage{amsfonts}
\usepackage{pgfplots}
\usepackage{tikz}
\usepackage{graphicx}
\usepackage[normalem]{ulem}

\usetikzlibrary{positioning}
\usetikzlibrary{shadows}
\usetikzlibrary{shapes.multipart}
\usetikzlibrary{fit}

\newcommand{\think}[1]{#1}

\definecolor{taborange}{RGB}{255, 127, 14}
\definecolor{tabgreen}{RGB}{44, 160, 44}
\definecolor{tabblue}{RGB}{31, 119, 180}

\newcommand*\de{\mathop{}\!\mathrm{d}}
\DeclareMathOperator*{\argmin}{arg\min}

\newcommand{\Aset}[0]{\mathcal{A}}
\newcommand{\Sset}[0]{\mathcal{S}}
\newcommand{\Xset}[0]{\mathcal{X}}
\newcommand{\xvec}[0]{\mathbf{x}}
\newcommand{\xivec}[0]{\bm{\xi}}

\newcommand{\bvec}[0]{\mathbf{b}}

\newcommand{\phivec}[0]{\bm{\phi}}
\newcommand{\omegavec}[0]{\bm{\omega}}

\newcommand{\state}[0]{s}
\newcommand{\action}[0]{a}

\newcommand{\reward}{\mathbf{r}}
\newcommand{\State}{S}
\newcommand{\Action}{A}

\newcommand{\Reward}{R}

\newtheorem{theorem}{Theorem}
\newtheorem{lemma}{Lemma}

\newtheorem{proposition}{Proposition}

\makeatletter
\newenvironment{customlegend}[1][]{%
	\begingroup
	\pgfplots@init@cleared@structures
	\pgfplotsset{#1}%
}{%
	\pgfplots@createlegend
	\endgroup
}%
\def\addlegendimage{\pgfplots@addlegendimage}
\makeatother
\newcommand\smalldots{\hbox to 1em{.\hss.\hss.}}

\begin{document}
%

%
\twocolumn[
\icmltitle{A Temporal-Difference Approach to Policy Gradient Estimation}
\icmlsetsymbol{equal}{*}
\begin{icmlauthorlist}
	\icmlauthor{Samuele Tosatto}{alberta}
	\icmlauthor{Andrew Patterson}{alberta}
	\icmlauthor{Martha White}{alberta,amii}
	\icmlauthor{A. Rupam Mahmood}{alberta,amii}
\end{icmlauthorlist}

\icmlaffiliation{alberta}{Department of Computer Science, University of Alberta, Edmonton, Canada}
\icmlaffiliation{amii}{CIFAR AI Chair, Alberta Machine Intelligence Institute (Amii)}

\icmlcorrespondingauthor{Samuele Tosatto}{tosatto@ualberta.ca}

\icmlkeywords{Policy Gradient, Temporal-Difference, Actor Critic, Gradient Critic}

\vskip 0.3in
]

\printAffiliationsAndNotice{\icmlEqualContribution}
\begin{abstract}
The policy gradient theorem (Sutton et al.,\ 2000) prescribes the usage of a cumulative discounted state distribution under the target policy to approximate the gradient.\
Most algorithms based on this theorem, in practice, break this assumption, introducing a distribution shift that can cause the convergence to poor solutions.\
In this paper, we propose a new approach of reconstructing the policy gradient from the start state without requiring a particular sampling strategy.\
The policy gradient calculation in this form can be simplified in terms of a \textsl{gradient critic}, which can be recursively estimated due to a new Bellman equation of gradients.\
By using temporal-difference updates of the gradient critic from an off-policy data stream, we develop the first estimator that side-steps the distribution shift issue in a model-free way.\
We prove that, under certain realizability conditions, our estimator is unbiased regardless of the sampling strategy.\
We empirically show that our technique achieves a superior bias-variance trade-off and performance in presence of off-policy samples.
The implementation of the experiments can be found at \href{github.com/SamuelePolimi/temporal-difference-gradient}{https://github.com/SamuelePolimi/temporal-difference-gradient}.
\end{abstract}
\section{Introduction}
Policy gradient methods provide an elegant approach to learn a parameterized policy in reinforcement learning \cite{deisenroth_survey_2013}. 
The policy gradient theorem \citep{sutton_policy_2000} provides a form for the gradient of the policy objective that can be sampled in a model-free way. 
This early work laid the foundation for practical methods, but as yet there is much more work to be done to provide effective approximations for the gradient. 
This is specially important in the off-policy setting, where we need the gradient under the current policy (the target policy) but the agent's experience is generated under a different policy (the behavior policy).
Addressing this gap is critical for building sample-efficient methods that permit re-use of the agent's experience either from past policies as replay \citep{mnih_human-level_2015,lillicrap_continuous_2016}, offline datasets \citep{levine_offline_2020}, or human demonstrations. 
%
The difficulty of constructing policy gradient approximators arises from the need to sample states from the discounted state-distribution that is induced by the target policy.
Although such sampling can be achieved in an on-policy manner, this approach is rarely used in practice, as 
it requires samples to be used only once, causing high variance and sample inefficiency.
Instead, most methods reuse data, introducing some   bias 
but compensating it with better sample efficiency.


An alternative approach consists in correcting the state distribution. 
The most straightforward choice is to use importance sampling to reweight states, as if they had been sampled proportionally to the target policy \citep{shelton_policy_2001,peshkin_learning_2002}. 
These methods are usually unbiased but affected by prohibitively large variance \citep{owen_monte_2013}. 
Many recent papers aim to lower the variance of pure importance sampling correction. \citet{liu_breaking_2018} and \citet{liu_off-policy_2019} introduce the concept of state-wise importance sampling. \citet{imani_off-policy_2018} proposes to combine semi-gradient with an emphatic weighting. Notably, they come across the gradient Bellman equation in their derivation. AlgaeDICE \citep{nachum_algaedice:_2019} incorporates a correction of the off-policy distribution by relying on the dual problem of a modified objective that incorporates an $f$-divergence regularization. But as yet more work is needed to make state-reweighting a practical choice. 

The most common choice has been to simply omit any correction to the state distribution 
such as is done in OffPAC \citep{degris_off-policy_2012}, DDPG \citep{lillicrap_continuous_2016}, A3C \citep{mnih_asynchronous_2016}, TD3 \citep{fujimoto_addressing_2018} and SAC \citep{haarnoja_soft_2018}. These methods are referred to as \textsl{semi-gradient} methods since the shift in the distribution can be seen as a result of the omission of a term in the gradient computation \citep{imani_off-policy_2018}. Though most state-of-the-art policy gradient algorithms use this semi-gradient approach, there are well known counterexamples showing that the bias can result in poor solutions, in both on-policy \citep{nota_is_2020,thomas_bias_2014} and off-policy settings \citep{imani_off-policy_2018,liu_off-policy_2019}. These counterexamples are not pathological and indicate issues that can arise under reasonable state aliasing. 
\cite{fujimoto_off-policy_2019} suggests that the effectiveness of aforementioned approaches can be hindered when the distribution shift is more pronounced.

In this work, we side-step the issue of the state weighting by pursuing an alternative form for the policy gradient. 
We propose learning a parametric representation of the cumulative discounted sequence of gradients generated from the target policy, which we call \textsl{the gradient critic}. The gradient critic can be learned from off-policy data using classic temporal-difference (TD) approaches. We will see that the gradient critic satisfies a Bellman equation (which we call \textsl{gradient Bellman equation}), allowing us to leverage the rich body of literature of value function estimators, including the ones for off-policy setting.
The gradient critic can be queried on the starting states, allowing us to predict the policy gradient without constraining ourselves to the need of a particular state distribution (i.e., the target-policy state-distribution).
We show that the gradient estimate is unbiased when the gradient function is realizable. To our knowledge, our method is the first to allow unbiased and model-free estimation of the policy gradient without using a state distribution reweighting\footnote{At the time of our submission, \citet{ni_optimal_2022} indipendently released on Arxiv a similar idea corroborating our findings.}.
\section{The Issue of State-Reweighting}
\label{sec:background}
We consider a Markov decision process $\mathcal{M} = (\!\Sset, \Aset, r,$  $p, \gamma, \mu_0\!)$, where $\Sset$ represents a finite set of states, $\Aset$ a finite set of actions\footnote{We provide a continuous state-action formulation in appendix.}, $r:\Sset\times\Aset\to\mathbb{R}$ a reward signal, $p(\state'|\state,\action)$ a probability density of transitioning to state $\state'$ after the application of action $\action$ in state $\state$, $\gamma$ is the discount factor, and $\mu_0$ is the distribution of the starting state.
The parameterized policy $\pi_\theta$, with parameters $\theta \in \mathbb{R}^{n_p}$, is a stochastic mapping, with density $\pi_\theta(\action|\state)$ over actions. We assume $\pi_\theta$ to be differentiable w.r.t.\ $\theta$.
We denote with $S_t, A_t, R_t$ the random variables representing the state, action, and reward at time $t$. We denote a sequence of state, action, and reward with $\tau_\pi$, when they are on-policy and with $\tau_\beta$ otherwise. 

The objective in the episodic setting is to maximize the expected discounted return from the start states
\begin{align*}
	J(\theta) &= {\tiny (1\!-\!\gamma)}\underset{\tau_\pi}{\mathbb{E}}\left[\sum_{t=0}^\infty \gamma^tR_t\right] 
\!	=\! (1\!-\!\gamma) \!\!\underset{\substack{S_0 \sim \mu_0\\
			 A_0 \sim \pi_\theta}}{\mathbb{E}}\left[Q^\pi(S_0, A_0) \right],
\end{align*}
where the action-values $Q^\pi(S, A)$ are the expected return under the policy from a given state and action, 
defined recursively as for all $\state \in \Sset$ and $\action \in \Aset$
\begin{equation}\small
	\think{Q^\pi(\state, \action) \!= \! r(\state, \action)\! +\! \gamma\! \sum_{\state',\action'}\! Q^\pi(\state'\!, \action')p(\state'|\state,\! \action)\pi_\theta(\action' | \state')}. \label{eq:bellman}
\end{equation}
\textbf{Policy Gradient Theorem.} One of the most important results in reinforcement learning is the policy gradient theorem (PGT), which allows us to estimate the policy gradient via samples:
%
%
\begin{align}
	\nabla_\theta J(\theta) = \underset{S \sim \mu_{\gamma}^\pi, A \sim \pi_\theta}{\mathbb{E}}\left[Q^{\pi_\theta}(S, A)\nabla_{\theta}\log\pi_{\theta}(A| S)\right]. \label{eq:pgt}
\end{align}
Note that this gradient has states sampled from the discounted state districution, i.e., $S \sim \mu_{\gamma}^\pi$, which is defined as follows.
The state-distribution $\mu^{\pi}_t(\state)\! =\! p_\pi(S_t\!=\! \state)$ indicates the density or the probability of the state $\state$ being observed at time $t$ when following $\pi$. The discounted state distribution is $\mu^{\pi}_\gamma(\state) = (1-\gamma)\sum_{t=0}^\infty\gamma^t\mu^{\pi}_t(s)$.

\textbf{Semi-Gradient.}
Most algorithms do not sample from the discounted state distribution $\mu_\gamma^\pi$, as they do not perform proper discounting \cite{nota_is_2019} and they reuse past experience collected in the replay buffer. The semi-gradient estimate can be seen as
\begin{align}
	\!\!\!\nabla_\theta^{S\!G} J(\theta) \!=\!\!\!\! \underset{\substack{S \sim \mu^\beta\\
			A \sim \beta}}{\mathbb{E}}\!\left[\tfrac{\pi_\theta(A|S)}{\beta(A|S)}Q^{\pi_\theta}(S, A)\nabla_{\theta}\log\pi_{\theta}(A| S)\right], \label{eq:sg}
\end{align}
where $\beta$ is a behavior policy, and $\mu_\beta$ its induced state-distribution.
Notice that the importance sampling in Equation~\eqref{eq:sg} only corrects the mistach in the action distribution but not the off-policy distribution $\mu_\beta$.
Examples of semi-gradient approaches are 
OffPAC \citep{degris_off-policy_2012}, DDPG \citep{silver_deterministic_2014} and SAC \citep{haarnoja_soft_2018,heess_learning_2015}. 

To avoid the semi-gradient problem, most approaches propose to perform a state-reweighting. The simplest version, proposed by \cite{shelton_policy_2001} and \cite{peshkin_learning_2002}, consists of multiplying all the importance sampling corrections along the trajectory, causing high variance. Recent work \cite{imani_off-policy_2018,liu_breaking_2018,liu_off-policy_2019} aims to lower the variance, but still relies on forms of importance sampling corrections. 
A proper and practical way of state-reweighting remains to be one of the critical issues for effective policy gradient estimation.
%
%
%
\section{Policy Gradient Using a Gradient Critic}\label{sec:gradient-function}
In this section, we pursue another path to estimating the policy gradient, by introducing the notion of a \textsl{gradient critic}. This gradient critic is the discounted accumulation of gradients, and as we show later, can be estimated using standard temporal-difference methods. 
This approach avoids the need to reweighting state distribution or incorporate high-variance importance sampling ratios, without incurring the high bias of semi-gradient approaches.

To obtain our alternative gradient estimator, we use a different formulation of the policy gradient theorem
\begin{align}
	\nabla_\theta J(\theta) \propto \mathbb{E}_{\tau_\pi}\Big[\sum_{t=0}^\infty \gamma^t \mathbf{g}_t\Big] \label{eq:pg-trajectory},
\end{align}
where $\mathbf{g}(S,A) = Q^\pi(S, A)\nabla_\theta \log \pi_\theta(A|S)$, $\mathbf{g}_t = \mathbf{g}(S_t,A_t)$\footnote{As discussed in Appendix~\ref{app:continuous-action}, the likelihood-ratio gradient (LR) in $\mathbf{g}$ can, in principle, be replaced with reparametrization gradient (RP) (similar to SAC \cite{haarnoja_soft_2018}), compositions of LR and RP \cite{lan_model-free_2022}, or others \cite{carvalho_empirical_2021}.}.
Equation~\eqref{eq:pg-trajectory} is equivalent to Equation~\eqref{eq:pgt}, with a constant factor $1-\gamma$ omitted. This form, however, highlights that the policy gradient can be seen as the discounted cumulation of gradients induced by \textsl{on-policy trajectories}.

We derive this form here, and connect it to what we call the gradient critic $\bm{\Gamma}^\pi(\state, \action) \doteq \nabla_\theta Q^\pi(\state, \action)$. Let us return to the definition of the objective $J$ and attempt to naively compute the gradient, using the chain rule,
	\begin{align}
			&\nabla_\theta J(\theta) \! \propto \! \sum_{\state, \action} \! \mu_0\!(\state)\!\big(\! \pi_\theta(\!\action | \state\!)  \nabla_\theta Q^{\pi_\theta}\!(\!\state,\! \action\!)\! +\!
		Q^{\pi_\theta}\!(\!\state, \! \action\!)\nabla_\theta\pi_\theta(\!\action |\state\!)\!\big) \nonumber\\
		& =  \sum_{\state, \action} \mu_0(\state) \pi_\theta(\action | \state)[ \mathbf{g}(\state, \action) +  \bm{\Gamma}^\pi(\state, \action)] \label{eq:first-derivation}
	\end{align}
We can derive a formula for $\nabla_\theta Q^\pi(\state, \action)$ by taking the derivative of the Bellman equation in Equation~\eqref{eq:bellman}
\begin{align}
\nabla_\theta Q^\pi(\state, \action)=&\gamma\sum_{\state', \action'}\Big( Q^\pi(\state', \action')\nabla_\theta\log\pi_\theta(\action'|\state') \nonumber   \\
	& + \nabla_\theta Q^\pi(\state', \action')\Big)\pi_\theta(\action'|\state')p(\state'|\state,\action). \nonumber
\end{align}
%
%
By substituting $\nabla_\theta Q^\pi$ with $\bm{\Gamma}^\pi$ and the integral with an expectation, we obtain the 
following recursive form
\begin{align}
\bm{\Gamma}^\pi\!(\state,\! \action\!)\!=\!\gamma\mathbb{E}[\mathbf{g}_{t+1}\! +\! \bm{\Gamma}^\pi\!(S_{t+1}, A_{t+1}\!)| S_t \!=\! \state,\! A_t \! =\! \action]. \label{eq:gbe}
\end{align}
We can unroll this recursion, expanding $\bm{\Gamma}^\pi(S_{t+1}, A_{t+1})$,
\begin{align}
	&\bm{\Gamma}^\pi(\state,\! \action)\!=\!\gamma\mathbb{E}[\mathbf{g}_{t+1}\! +\! \bm{\Gamma}^\pi(S_{t+1}, A_{t+1})| S_t \!=\! \state,\! A_t \! =\! \action]\nonumber  \\
	&= \gamma\mathbb{E}[\mathbf{g}_{t+1} + \gamma \mathbf{g}_{t+2} + \gamma \bm{\Gamma}^\pi(S_{t+2}, A_{t+2})| S_t\! =\! \state,\! A_t\! =\! \action] \nonumber\\
	&= \mathbb{E}\bigg[\sum_{t=1}^{\infty} \gamma^t\mathbf{g}_t | S_t = \state, A_t = \action\bigg]. \label{eq:unrolled-gradient-critic}
	\end{align}
From Equations~\eqref{eq:first-derivation} and \eqref{eq:unrolled-gradient-critic} we can verify that 
\begin{align}
 \nabla_\theta J(\theta)\! \propto\! \underset{\substack{S \sim \mu_0 \\A \sim \pi_\theta}}{\mathbb{E}}
\big[\mathbf{g}(S,A) \!+\! \bm{\Gamma}^\pi(S, A)\big]. \label{eq:gradient-critic-starting}
\end{align}
Note that the gradient critic can recover the policy gradient just by computing an expectation over the starting-state distribution. In other words, given an estimated gradient critic $\hat{\bm{\Gamma}}^\pi$, an estimated value critic $\hat{Q}^\pi$, and a start state $\state_0$, the policy can be updated by using $a^\pi_0 \sim \pi_\theta(\cdot |s_0)$ and
\begin{align}
\theta \gets \theta + \eta [\hat{\mathbf{g}}(\state_0, \action^\pi_0) + \hat{\bm{\Gamma}}^\pi(\state_0, \action_0^\pi)], \label{eq:gradient-ascent}
\end{align}
where $\hat{\mathbf{g}}(\state_0, \action^\pi_0) = \hat Q^\pi(\state_0, \action^\pi_0)\nabla_\theta \log \pi_\theta(\action^\pi_0|\state_0)$. This policy gradient estimator is  naturally model-free and off-policy, does not require state distribution reweighting, and has less variance than the classic policy gradient, as it involves overall less stochasticity. 

There is, of course, a big caveat: we require an estimate of this gradient critic. Poor estimates may introduce significant bias, overriding the benefits of this variance reduction. Even worse, we compound two approximations: an approximate value critic $\hat{Q}^\pi$ and gradient critic $\hat{\bm{\Gamma}}^\pi$. Remarkably, we find that we can actually obtain an unbiased gradient estimate, under linear function approximation with realizability, using a (batch) TD approach for learning both the value critic and gradient critic. We prove this later, in Theorem \ref{theo:perfect-features}, after introducing the gradient critic estimation approaches. This theorem is particularly surprising because semi-gradient approaches remain biased, even with the knowledge of a perfect critic \citep{imani_off-policy_2018,liu_off-policy_2019}. 

Aside the independent work of \citet{ni_optimal_2022}, to the best of our knowledge, this is the first unbiased policy gradient approach estimator, with function approximation, that does not rely on state distribution reweighting. Notably, \citet{tosatto_nonparametric_2020,tosatto_batch_2021} derived a similar approach to Equation~\eqref{eq:gradient-ascent} based on nonparametric statistics. Their method, however, do not scale with samples and requires infinitesimal bandwidth of the kernels to ensure unbiasedness. 

In practice, of course, we may not have realizability and we need to understand when this approach will succeed and when it will fail. In the remainder of this paper, we investigate the properties of this approach, particularly focusing on different estimation approaches for the gradient critic and assessing those approaches empirically. 
\section{Estimating the Gradient Critic}
\label{sec:estimating-critic}
In this section, we discuss the gradient Bellman equation and how we can use it to estimate the gradient critic $\bm{\Gamma}^\pi$. Notice that $\bm{\Gamma}^\pi(\state, \action) = \nabla_\theta Q^\pi(\state, \action)$ represents the differentiation of the state-action value with respect to the policy's parameters, is different from reparameterization gradient, and cannot, in general, be found in closed form or via automatic differentiation. Instead, we leverage the gradient Bellman equation and explain how to use the a TD algorithm to estimate this gradient critic. The basic idea is that the gradient estimator can be used with Equation~\eqref{eq:gradient-ascent} to update the policy gradient. Later in Section \ref{sec_extensions}, we outline an $n$-step estimator that is robust to the bias of the gradient critic.
%
%
\subsection{The Gradient Bellman Equation}
The \emph{gradient Bellman equation} was already shown in Equation \eqref{eq:gbe}, though we had not yet given it a name.
The equation is $\bm{\Gamma}^\pi(\state, \action)=\gamma\mathbb{E}[\mathbf{g}_{t+1} + \bm{\Gamma}^\pi(S_{t+1}, A_{t+1})| S_t = \state, A_t = \action] $. 
For the $i$-th element $g_{i}$ of $\mathbf{g}$, we have that
\begin{align}
	\Gamma^\pi_i(\state, \action) \!= \! \sum_{\state',\action'} \! \left(g_i( \state', \action') \! +\! \gamma  \Gamma^\pi_i(\state', \action')\pi(\action'|\state')\right)\! p(\state'|\state,\!\action), \nonumber 
\end{align}
which is a Bellman equation for a scalar $\Gamma^\pi_i(\state, \action)$. Therefore, Equation~\eqref{eq:gbe} is a Bellman equation for the vector $\bm{\Gamma}^\pi(\state, \action)$, composed of this set of independent Bellman equations. 
That is why we call \eqref{eq:gbe} the \textsl{gradient Bellman equation}.

Bellman equations are well studied, giving us broad literature about approximation techniques and theoretical results. For example, $\bm{\Gamma}^\pi(\state, \action)$ can be estimated using bootstrapping approaches, like temporal-difference learning. One key subtlety is that the term $g_i(\state, \action)$ involves $Q^\pi$, which also needs to be estimated. Fortunately, we already estimate this term, the standard value critic, in actor-critic methods.
\subsection{An Online Estimator using TDRC}\label{sec_td}
\newcommand{\gdelta}{\boldsymbol{\delta}^{g}}
\newcommand{\gparams}{\mathbf{G}}
\newcommand{\gcritic}{\hat{\bm\Gamma}}
The full algorithm involves 1) estimating a value critic, 2) using the value critic to estimate the gradient critic, and 3) using both the value and the gradient critics to estimate the policy gradient update. In this section, we explain an algorithm based on TD with regularized correction \cite{ghiassian_gradient_2020}, and detailed in Algorithm~\ref{alg:tdrc-gamma}.

We can use TD to estimate both the standard action-value critic, as well as the gradient critic. We now have two temporal-difference errors: $\delta_t$ for the value critic, and the vector $\gdelta_t$ for the gradient critic, which is the size of the number of policy parameters. Let $ \omegavec_t$ be the parameters for the value critic $\hat{Q}_t$ and $\gparams_t$ the parameters for the gradient critic $\gcritic$. The updates for the value critic, with step-size $\alpha_t$, are the standard TD updates:
	\begin{align}
		 \delta_t &= \Reward_{t+1} + \gamma \hat{Q}_t(\State_{t+1}, \Action_{t+1}) - \hat{Q}_t(\State_t, \Action_t), \nonumber \\
		\omegavec_{t+1}  &= \omegavec_t + \alpha_t \delta_t \nabla_{\omegavec_t} \hat{Q}_t(\State_t, \Action_t). \label{eq_tdvalue_critic}
	\end{align}
	Similarly, we can get vector-valued TD updates for the gradient critic as follows:
			\begin{align}
		\mathbf{g}_t &= \hat{Q}_t(\State_t, \Action_t^\pi) \nabla_\theta\log\pi_\theta(\Action_t^\pi|\State_t),\nonumber\\
		\gdelta_t &= \mathbf{g}_{{t+1}} + \gamma \gcritic(\State_{t+1}, \Action_{t+1}) - \gcritic(\State_t, \Action_t), \nonumber \\
		\gparams_{t+1} &= \gparams_t + \alpha_t \gdelta_t \nabla_{\mathbf{G}_t} \gcritic(\State_t, \Action_t).  \label{eq_tdgradient_critic}
	\end{align}
%
Simple TD techniques \citep{sutton_learning_1988} are sample efficient but are not guaranteed to converge with off-policy data \citep{baird_residual_1995}.
Gradient TD methods \citep{sutton_convergent_2008} and TD with gradient corrections (TDC) \citep{sutton_fast_2009} are guaranteed to converge under general conditions, however are often less sample efficient than TD \citep{ghiassian_gradient_2020}.
A recent approach called TDRC (temporal-difference with regularized correction, Ghiassan et al., \yrcite{ghiassian_gradient_2020}) proposes to mix regular TD with TDC, allowing convergence with off-policy samples without losing sample efficiency.

Once the gradient critic has been estimated, it can be used to update the policy parameters as in Equation~\eqref{eq:gradient-ascent}. Algorithm~\ref{alg:tdrc-gamma} details a pseudocode of TDRC with policy improvement.

\section{Unbiased Estimation Under Realizability}
\label{sec:unbiased}
In this section we analyze the properties of the gradient critic, obtained with TD under linear function approximation. In particular, we show that the gradient critic given by the TD fixed-point solution in the realizable setting---the case where the features are sufficient to represent the value critic---gives an unbiased estimate of the policy gradient.

The TD fixed-point solution of the projected Bellman equation induced by \eqref{eq:gbe} is as follows:
	\begin{align}
		& \hat{\bm{\Gamma}}_{T\!D\!Q}^\pi(\state, \action) = \phivec^\intercal(\state, \action)\mathbf{G}_{T\!D\!Q} \ \text{with} \ \mathbf{G}_{T\!D\!Q} = \mathbf{A}_\pi^{-1}\mathbf{B}_{Q},  \nonumber \\
		& \mathbf{A}_{\pi} = \mathbb{E}_{\zeta}\left[\phivec(S, A)\left(\phivec^\intercal(S, A) - \gamma\phivec^\intercal(S', A')\right)\right], \nonumber\\
		& \mathbf{B}_Q = \gamma \mathbb{E}_\zeta\left[\phivec(S, A)Q^\pi(S', A')\nabla_\theta \log\pi_\theta(A'|S')\right]. \label{eq:a-b-first}
	\end{align}
We first consider approximation error of the gradient critic assuming access to the true value critic $Q^\pi$.
\begin{lemma}[Gradient Critic with Perfect Value Critic]\label{lemma:tdq}
		Let us consider finite state and action sets. For an irreducible Markov chain induced jointly by the transition function $p$ and the policy $\pi_\theta$ having steady distribution $\mu$. Let $\zeta$ be a process where $S \sim \mu_\beta$, $A \sim \beta(\cdot | S)$, $S' \sim p(S' |S, A)$ and $A' \sim \pi_\theta(\cdot |S')$. 	If $p(S, A) = \mu_\beta(S)\beta(A|S)$ satisfies the inequality introduced by \cite{kolter_fixed_2011}, then
	\begin{align}
		& \| \hat{\bm{\Gamma}}^\pi_{T\!D\!Q}(\state, \action) - \nabla_\theta Q^\pi(\state, \action)\|_{\zeta} \leq \nonumber \\
		&\frac{1 + \kappa \gamma}{1-\gamma} \min_{\mathbf{G}}\left\|\phivec^\intercal(\state,\action)\mathbf{G} - \nabla_\theta Q^\pi(\state, \action)\right\|_{\zeta},
	\end{align}
	with  $\kappa\! =\! \max_{\state, \action}h(\state, \action)/\min_{\state,\action} h(\state, \action)$ where $ h(\state, \action) = \sqrt{\mu(\state)\pi_\theta(\action|\state)}/\sqrt{\mu_\beta(\state)\beta(\action|\state))}$.
\end{lemma}

Next we consider the more realistic setting where we estimate the value critic.
\think{Again, because we use TD methods, we will use the TD-fixed point solution $\hat{Q}_{T\!D}^\pi(\state, \action) = \bm{\varphi}^\intercal(\state, \action)\omegavec_{T\!D}$. Namely, we have
$\omegavec_{T\!D} = \mathbf{C}^{-1}_{\pi}\mathbf{b}$ with
\begin{align}
&	\mathbf{C}_{\pi}= \mathbb{E}_{\zeta}\left[\bm{\varphi}(S, A)\left(\bm{\varphi}^\intercal(S, A) - \gamma\bm{\varphi}^\intercal(S', A')\right)\right],\nonumber \\ 
&	\mathbf{b}  = \mathbb{E}_\zeta[\bm{\varphi}(S, A)r(S, A)], \ \text{and} \nonumber \\
&	\hat{\bm{\Gamma}}_{T\!D}^\pi(\state,\action)\! =\!  \phivec^\intercal(\state, \action) \mathbf{G}_{T\!D}, \; \text{with}\; \mathbf{G}_{T\!D}\! =\! \mathbf{A}_\pi^{-1}\mathbf{B}, \label{eq:gradient-td}
\end{align}
where $\mathbf{B}  = \gamma \mathbb{E}_\zeta[\phi(S, A)\hat{Q}^\pi_{T\!D}(S', A')\nabla_\theta\log\pi_\theta(A'|S')]$ is different from $\mathbf{B}_Q$ in (\ref{eq:a-b-first}) and $\mathbf{A}_{\pi}$ is as before.}

The approximation error of the gradient critic when using the TD fixed-point solution for approximating both the critic can be bounded.
\begin{theorem}[Error Analysis]\label{theo:GLS}
	Consider the assumption in Lemma~\ref{lemma:tdq} and Proposition~\ref{prop:generalized-ls} (Appendix \ref{app:generalized-least-squares}). The TD fixed-point solution $\hat{\bm{\Gamma}}_{T\!D}^\pi$  of the gradient function defined in Equation~\ref{eq:gradient-td} satisfies
	\begin{align}
		&\|\hat{\bm{\Gamma}}^\pi_{T\!D}(\state, \action) - \nabla_\theta Q^\pi(\state, \action)\|_{\zeta} \leq \nonumber \\
		&\frac{1 + \gamma\kappa}{1-\gamma} \min_{\mathbf{G}}\left\|\phivec^\intercal(\state,\action)\mathbf{G} - \nabla_\theta Q^\pi(\state, \action)\right\|_{\zeta} + \nonumber \\
		& \gamma n_p b \kappa \frac{(1+\gamma\kappa)^2}{(1-\gamma)^2}\min_{\omegavec}\|\bm{\varphi}^\intercal(\state,\action)\omegavec - Q^\pi(\state,\action)\|_{\zeta}, \nonumber
	\end{align}
	with $b = | \max_{\action, \state, i} \partial/\partial\theta_i\log\pi(\action|\state)|$, $n_p$ the number of the policy's parameters  and $\kappa$ as in Lemma~\ref{lemma:tdq}.
\end{theorem}
The proof of Theorem~\ref{theo:GLS} relies on the results in \citep{kolter_fixed_2011} and on Lemma~\ref{lemma:gradient-td} (Appendix~\ref{app:least-squares-gradient}).

\textbf{Remark:} Theorem~\ref{theo:GLS} shows that the approximation error of the TD fixed-point solution of the gradient critics is bounded by the projection error of both critics: if the feature spaces of both the value and the gradient critic are good enough, both the projection errors goes to zero, ensuring an unbiased gradient estimate.

\subsection{Shared Features}
The gradient critic is inherently more complex than the value critic, since it predicts a high-dimensional quantity. It seems resonable that the feature space should also be larger (w.r.t. the value critic's one) to compensate this complexity.
Surprisingly, in this linear setting, when the feature space of the value critic allows an unbiased value estimate, then it can be also reused by the gradient critic to obtain an unbiased gradient estimate.

Consider sharing the features between value and gradient critic, i.e., $\bm{\phi} = \bm{\varphi}$. Notice that, in this case, $\hat{Q}_{T\!D}^\pi(\state, \action) = \bm{\phi}^\intercal(\state, \action) \omegavec_{T\!D}$, and $\omegavec_{TD} = \mathbf{A}_\pi^{-1}\mathbf{b}$ where $\mathbf{b} = \mathbb{E}_\zeta[\bm{\phi}(S, A)r(S, A)]$.

In this case, it is possible to show that the TD fixed-point solution of the gradient critic is the gradient of TD fixed-point solution of the value critic.

\begin{lemma}\label{lemma:gradient-td}
	When $\bm{\phi} = \bm{\varphi}$, the gradient approximation $\hat{\bm{\Gamma}}_{T\!D}^\pi(\state,\action) $ and the TD fixed-point critic $\hat{Q}^\pi_{T\!D}(\state,\action)$ satisfies
	\begin{align}
		\hat{\bm{\Gamma}}^\pi_{T\!D}(\state, \action) = \nabla_\theta \hat{Q}^\pi_{T\!D}(\state, \action) \tag*{$\forall \state \in \Sset, \forall \action \in \Aset$.}
	\end{align}
\end{lemma}
\begin{proof} Consider $\nabla_\theta \hat{Q}^\pi_{T\!D}(\state, \action) = \phivec^\intercal(\state, \action)\nabla_\theta \omegavec_{T\!D}$ and
	\begin{align}
		& \nabla_\theta\omegavec_{TD} = - \mathbf{A}_{\pi}^{-1}
		\left(\nabla_\theta\mathbf{A}_{\pi}\right)
		\mathbf{A}_{\pi}^{-1} \mathbf{b} \nonumber  \\
		&= - \mathbf{A}_{\pi}^{-1}
		\left(\nabla_\theta \mathbf{A}_{\pi}\right)
		\omegavec_{TD}\nonumber  \\
		& = \gamma \mathbf{A}_\pi^{-1}\mathbb{E}_\zeta\left[\phivec(S, A)\phivec^\intercal(S', A')\omegavec_{TD}\nabla_\theta \log \pi_\theta(A' | S')\right]\nonumber  \\
		&=  \gamma \mathbf{A}_\pi^{-1}\mathbb{E}_\zeta\bigg[\phivec(S, A) \hat{Q}^\pi_{T\!D}(S', A') \nabla_\theta \log \pi_\theta(A' | S')\bigg].\nonumber \\
		&=  \mathbf{A}_\pi^{-1}\mathbf{B} = \mathbf{G}_{T\!D},\nonumber
	\end{align}
	implying $\nabla_\theta \hat{Q}^\pi_{T\!D}(\state, \action) =\phivec^\intercal(\state, \action)\mathbf{G}_{T\!D} = \hat{\bm{\Gamma}}^\pi_{T\!D}(\state, \action)$.
\end{proof}
This identity shows that the gradient predicted by the gradient critic is \textsl{consistent} with the gradient of the value critic.
Usually, policy gradient algorithms do not guarantee this consistency. In fact, after the policy update, the policy might improve, but this improvement might be not representable by the critic, causing instability. This issue is well known in value iteration, called \textsl{delusional bias} \citep{lu_non-delusional_2018}. When features are shared, the converged gradient critic predicts the gradient of \textsl{the approximated value critic} w.r.t.\ the policy parameters, guaranteeing its improvement.

The benefit of sharing features is emphasized in  Theorem~\ref{theo:perfect-features}, where the realizability of the value critic implies the realizability of the gradient critic.
%
\begin{theorem}[Perfect Features]\label{theo:perfect-features} Let $\Phi \equiv \{\phi(\state, \action)  | \forall \state \in \Sset \land \action \in \Aset\}$ and  $\Phi' \equiv \{\phi(\state, \action) - \gamma \sum_{\state', \action'} \phi(\state', \action')\pi_\theta(\action'|\state)p(\state'|\state, \action) | \forall \state \in \Sset \land \action \in \Aset\}$ be $n_f$-dimensional \think{vector spaces} (they both admit at least one basis of dimension $d_f$). Let $\mu_\beta$ be such that $\mathbf{A}_\pi$ is invertible. If we assume that for any policy parameter $\theta$ exists a parameter $\omegavec_\pi$ such that
	\begin{flalign}
		&& \phivec^\intercal(\state, \action)\omegavec_\pi &= Q^\pi(\state, \action)  && \forall \state \in \Sset \land \action \in \Aset, \nonumber \\
		\text{then} && \quad \hat{\bm{\Gamma}}_{T\!D}^\pi(\state, \action) &= \nabla_\theta Q^\pi(\state, \action) && \forall \state \in \Sset \land \action \in \Aset. \nonumber
	\end{flalign}
\end{theorem}
The proof can be found in Appendix~\ref{app:perfect-features}.
This theorem further empasizes that the gradient critic can be unbiased. In particular, even though the gradient critic predicts a higher-dimensional vector  compared to the value critic, it can still achieve a good approximation with the features used by the classic value critic.

%
%
\section{Controlling the Bias and Variance}\label{sec_extensions}
The proposed estimator fully relies on the gradient critic. We can instead reduce this reliance, by incorporating sampled gradient components and then bootstrapping. 

It is important to notice that Equations~\ref{eq:pg-trajectory} and \ref{eq:gradient-critic-starting} represent two extremes: the first is a full Monte-Carlo rollout, while the second uses full bootstrapping. By applying recursively the definition of the gradient critic, as we have done in the derivation of Equation~\ref{eq:unrolled-gradient-critic}, we can rewrite the policy gradient as a $n$-step estimator:
\begin{align*}
	\nabla_\theta J(\theta) \propto &
	\underset{\tau_\pi}{\mathbb{E}}\bigg[\sum_{t=0}^{n}\gamma^t\mathbf{g}_t + \gamma^{n}\bm{\Gamma}^\pi(S_{n}, A_{n}^\pi)\bigg].
\end{align*}
The advantage of this approach is that we can either immediately bootstrap off of our estimate of the gradient ($n = 0$), or we can wait one step to bootstrap, or we can wait $n$ steps.
This perspective highlights even more the role of $\bm{\Gamma}^\pi(\state, \action)$  as a critic function.

We can also express the $n$-step estimator under off-policy sampling. We can do so with the standard strategy of path-wise importance sampling corrections \cite{shelton_policy_2001}. Under behavior policy $\beta$, it yields the following
\begin{align*}
	\nabla_\theta J(\theta) \propto &
   \underset{\tau_\beta}{\mathbb{E}}\bigg[\sum_{t=0}^{n}\gamma^t\rho_t \mathbf{g}_t + \gamma^{n}\rho_n\bm{\Gamma}^\pi(S_{n}, A_{n}^\pi)\bigg],
\end{align*}
where $\rho_0 = 1$, $\rho_t = \prod_{i=0}^{t-1} \pi(A_i |S_i)/\beta(A_i | S_i)$, $\tau_\beta$ are off-policy trajectories, $\mathbf{g}_t = \mathbf{g}(S_t, A^\pi_t)$ and  $A_{n}^\pi \sim \pi_\theta(\cdot | S_n)$ are on-policy actions. These actions are sampled on-policy after $n$ steps and used to reduce the gradient's variance.

The utility of the $n$-step form for the PG is twofold: (a) it allows us to trade off bias and variance in our PG estimator, and (b) it allows us to mitigate the role of the state reweighting and the associated variance issues. Using $n = 1$ means that we rely heavily on our gradient critic, which might be biased. However, we avoid the variance of sequences of sampled $\mathbf{g}_t$, which we have for larger $n$.
This effect is pronounced in the off-policy setting, where for $n > 1$, we correct the whole trajectory distribution. As $n$ gets larger, we approach the classic PG estimator, with state reweighting given by the products of importance sampling ratios and $\gamma$. Therefore, the $n$-step estimator allows us to reduce the variance due both to sampled $\mathbf{g}_t$ and state reweighting. In the extreme, at $n = 0$, we do not need to use any reweighting, because $\bm{\Gamma}^\pi(\state, \action)$ allows us to query the gradient from any state and action.

Once we have this $n$-step estimator, it is straightforward to extend it to eligibility traces (Appendix~\ref{app:eligibility-trace}),
\begin{equation*}
	\nabla_\theta J(\theta) \propto  \underset{\tau_\beta}{\mathbb{E}}\bigg[\sum_{t=0}^{\infty}\lambda^t\gamma^t\rho_t\big( \mathbf{g}_t + (1-\lambda) \bm{\Gamma}^\pi(S_t, A_t^\pi)\big)\bigg],
\end{equation*}
with trace parameter $\lambda \in [0,1]$.
With $\lambda=0$, we obtain the $n = 0$ estimator, where we immediately bootstrap off of $\bm{\Gamma}^\pi$. As $\lambda\to 1$, we recover the classic PGT. This trace gradient is actually an exponential average, with weighting $\lambda$, of all $n$-step estimators, and so provides a smoother trade-off between bias and variance.

Finally, we can further reduce variance, at the cost of bias, by considering the PG without any state reweighting. Namely, we can instead blend between the semi-gradient and our approach, rather than the corrected gradient and our approach,
\begin{align}
	\!\!\!\nabla_\theta J(\theta) \approx &  \underset{\tau_\beta}{\mathbb{E}}\bigg[\sum_{t=0}^{\infty}\lambda^t\gamma^t\big( \mathbf{g}_t + (1-\lambda) \bm{\Gamma}^\pi(S_t,A_t^\pi)\big)\bigg]. \label{eq:semi-traces}
\end{align}
As $\lambda \rightarrow 1$, we recover the semi-gradient, because we are effectively sampling $S \sim \mu_{\gamma}^\beta$.
The bias-variance trade-off in this estimator is more subtle. For larger $\lambda$, we are more robust to bias in the gradient critic, but also suffer more from bias due to the omission of the importance sampling ratios. Therefore, when the gradient critic is quite accurate, a lower $\lambda$ might result in less bias. We showed in Theorem~\ref{theo:perfect-features} that in some cases, the gradient critic can be unbiased, even when estimated under off-policy samples. This result highlights that this generalized estimator can provide improvements on the classic gradient estimation, allowing us to avoid reweighting and potentially reducing the bias significantly. Algorithm~\ref{alg:tdrc-gamma} depicts a policy improvement scheme unifying the gradient critic estimate presented in Section~\ref{sec_td} with this extension to eligibility traces.
%
%
%
%
%
\begin{algorithm}[t]
	\caption{TDRC$\bm{\Gamma}$\label{alg:tdrc-gamma}}
	\begin{algorithmic}[1]
		\STATE \textbf{Input:} Set of features $\phi$, policy $\pi_\theta$, learning rates $\alpha_t$ and $\eta$, TDRC regularization factor $\beta$, eligibility trace $\lambda$, initial parameters $\omegavec_0$ and $\mathbf{G}_0$
		\STATE $\nu_0 = 1$,  $\state_0 \sim \mu_0$
		\FOR {$t=0$ to $T-1$:}
		\STATE Apply  $\action_t \sim \beta(\cdot | \state_t)$ on the environment
		\STATE Observe state $\state_t$ and reward $\reward_{t+1}$
		\STATE Draw actions $\action_t^\pi \sim  \pi_\theta(\cdot | \state_t)$, $\action_{t+1}^\pi \sim \pi_\theta(\cdot | \state_{t+1})$
		\STATE $\hat{Q}_t\!=\!\phivec^\intercal(\state_t, \action_t^\pi)\omegavec_t$, $\hat{\bm\Gamma}_t\! =\!\phivec^\intercal(\state_t, \action_t^\pi)\mathbf{G}_t$
		\STATE $\theta \! \leftarrow \!\theta \! +\! \eta \nu_t\left( Q_t\!\nabla_\theta\!\log\!\pi_\theta(\action_t^\pi | \state_t )\! + \!(1\!-\!\lambda)\hat{\bm\Gamma}_t\right)$
		\STATE Compute $\omegavec_{t+1}$, 
		and $\mathbf{G}_{t+1}$ 
		using TDRC (see Appendix~\ref{app:tdrc-gamma})
		\IF {$\state_t'$ is a terminal state:}
		\STATE $\nu_{t+1} = 1$,  $\state_{t+1} \sim \mu_0$
		\ELSE
		\STATE $\nu_{t+1} = \lambda\gamma\nu_t$, $\state_{t+1} = \state_t$
		\ENDIF
		\ENDFOR
	\end{algorithmic}
\end{algorithm}
\begin{figure*}[ht]
	\vspace{0.1cm}
	\includegraphics{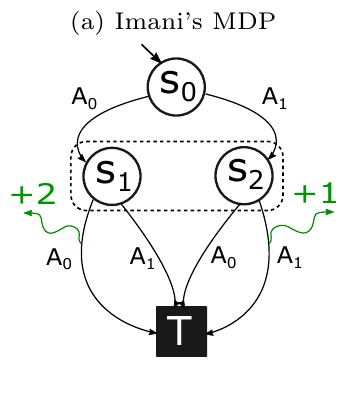}
	\includegraphics{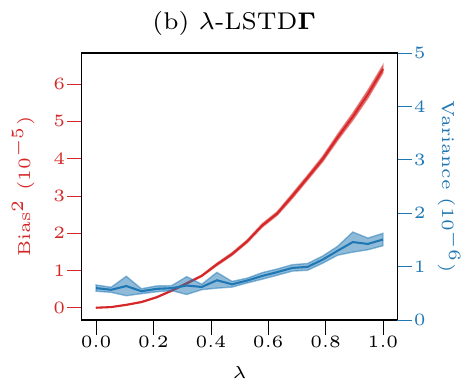}
	\includegraphics{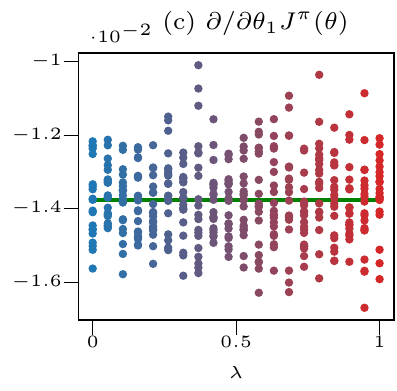}
	\hspace{0.3cm}
	\includegraphics{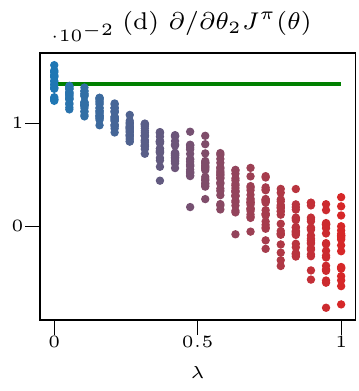}
	\vspace{-1.5em}
	\caption{(a) \texttt{Imani's MDP} \citep{imani_off-policy_2018}. (b) Bias and variance of gradient evaluation with LSTD$\bm{\Gamma}$ in \texttt{Imani's MDP}. Lower $\lambda$ achieves lower bias and variance, showing that the gradient critic helps delivering a better estimate. (c, d)  the scatter plots show single estimates of LSTD$\bm\Gamma$ and  green lines the ground truth. While low $\lambda$ helps the estimate of $\partial/\partial\theta_2$  it does not improve $\partial/\partial\theta_1$, suggesting that the gradient critic used on a convenient subset of parameters could still be beneficial. \label{fig:imani-bias}}
\end{figure*}
\section{Extension to Deep Reinforcement Learning}

The primary goal of this paper is to introduce the theoretical foundations of gradient critic algorithms.
To this end, we focused on linear function approximation; however, the concepts presented in can be extended to function approximation with deep neural networks.
The primary challenge is that the gradient critic estimates a vector of size $d$---the number of parameters in the neural network---resulting in a very large output.
To overcome this issue, we propose that the gradient critic can learn only a subset of the gradient, while still achieving a favorable bias-variance trade-off.

In fact, past literature---in addition to our own experiments---suggests that some gradients are more susceptible than others to distribution shift. \citet{imani_off-policy_2018} show that the distribution shift becomes detrimental when united with state aliasing.
When analyzing the estimation bias of the semi-gradient approach on their toy MDP (Figure 2a), we find that the gradient update was biased mainly for the parameters responsible for selecting the action corresponding to the aliased states (Section~\ref{fig:imani-bias}c, d, details in Appendix~\ref{app:bias-variance}).
In MDPs, the state is usually fully informative
but as information flows from the bottom to the top layers of the neural network, the learned features may introduce state aliasing in higher-level of abstraction.
We 
argue that learning the gradient of the last layer of the actor network will
potentially trade off the complexity of learning of a high-dimensional gradient, with the benefit introduced by our approach.
\section{Empirical Analysis}
We want to show that 1) the semi-gradient is generally biased, whereas the gradient critic is unbiased provided realizability, 2) this unbiasedness helps the convergence to better solutions, and 3) even when applied to a subset of the actor parameters, the gradient critic helps to attain higher performance.
We test four different algorithms: a classic semi-gradient algorithm,  OffPAC \citep{degris_off-policy_2012}, an actor critic algorithm with full importance sampling correction ACE(1) \citep{graves_off-policy_2021}, a simple and new policy gradient scheme called LSTD$\bm\Gamma$ that uses the least-squares temporal-difference solution for the gradient critic computed from offline data using Equations~\ref{eq:a-b-first}, and TDRC$\bm{\Gamma}$ as described in Algorithm~\ref{alg:tdrc-gamma}. 

\texttt{Imani's MDP} (Figure~\ref{fig:imani-bias}a) is designed to show the fallacy of semi-gradient methods under off-policy distribution. In their work, Imani et al.\ assumed a perfect critic but aliased states for the actor. 
In agreement with their setup, we use a behavior policy that samples with probability $0.25$ action $A_0$ and $0.75$ action $A_1$. The critic's features have sufficient information for all state-action pairs $\phivec(s, a) = \texttt{one-hot-encode}(s, a)$. The optimization policy is initialized with probabilities $0.9$ and $0.1$ for actions $A_0$ and $A_1$ respectively.\footnote{Differently from their setup, we measure the performance by using the return in Section~\ref{sec:background}, instead of their proposed off-policy objective. Furthermore, our returns are discounted by $1-\gamma$.}
\\
\texttt{Randomly Generated MDPs.} The MDP mentioned above is designed appositely to show the flaws of semi-gradient algorithms, and it assumes fully informative critic features. We want to test the gradient function in a more generic setting. To this end, we randomly generate $2500$ MDPs with $30$ states and $2$ actions. We use this task to study the effect of the application of the gradient critic restricted to a subset of the parameters.
\begin{figure*}[t]
	\fbox{
		\hspace{-0.3cm}
		\includegraphics{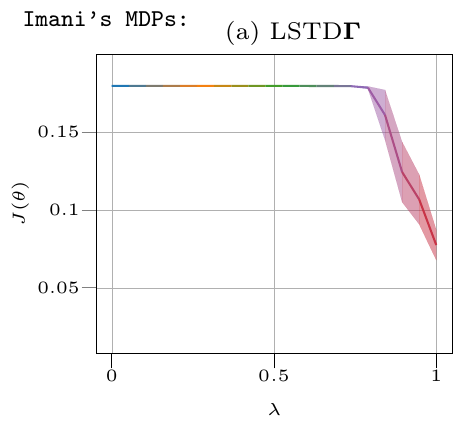}
		\hspace{-0.3cm}
		\includegraphics{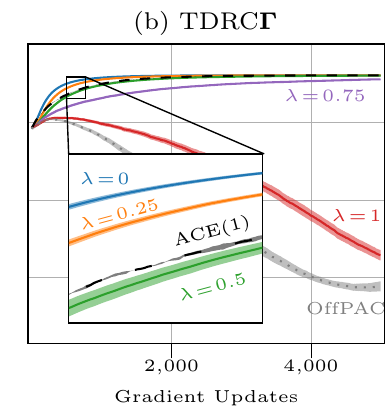}
		\hspace{-0.1cm}
	}
	\fbox{
		\hspace{-0.3cm}
		\includegraphics{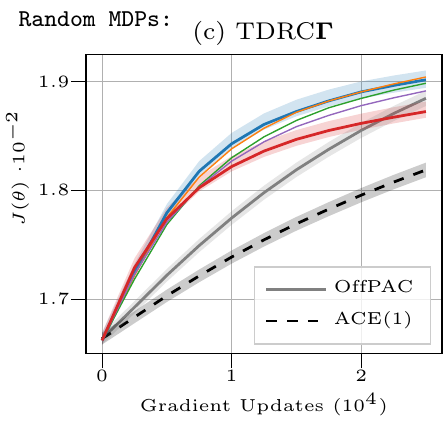}
		\hspace{-0.3cm}
		\includegraphics{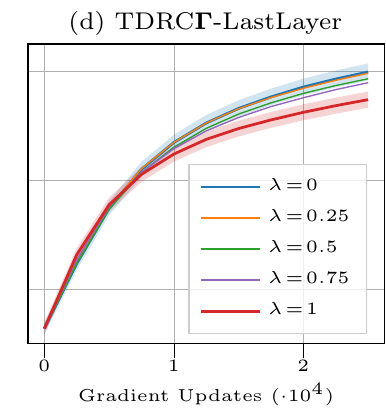}
		\hspace{-0.1cm}
	}
	\vspace{-1.1em}
	\caption{\texttt{Imani's MPD}: (a) final performance of LSTD$\bm\Gamma$ and (b) learning curve of TDRC$\bm{\Gamma}$. \texttt{Random MPDs}: (c) learning curve of classic TDRC$\bm\Gamma$ compared with OffPAC and ACE(1) (d) learning curves of TDRC$\bm\Gamma$ with the gradient critic applied only to the last layer of the actor. We notice that lower values of $\lambda$ improves the performance in both the tasks. Using the critic only on the last layer does not degrade the performance sensibly. Shaded areas show the standard error.  \label{fig:random-mdps}\label{fig:learning-curves}}
\end{figure*}

\subsection{Analysis on Imani's MDP}
The goal of this set of experiments is to analyze the effect of the gradient critic estimator on \texttt{Imani's MDP}. In particular we analyze the estimation bias, and variance and performance of our method for different value of $\lambda$.\\
\textbf{Bias-Variance Analysis.} We generate datasets of $500$ samples beforehand, using the behavioral policy. The target policy's parameters are initialized to match the condition described earlier. We estimate the gradient using LSTD$\bm{\Gamma}$ with $\lambda \in \{0, 0.05, 0.1, \dots, 1\}$. For each value of $\lambda$, we compute $1000$ estimates of the bias and the variance accompanied with confidence intervals at $95\%$. Figure~\ref{fig:imani-bias}b shows that both the bias and the variance of the estimator increase as $\lambda$ increases. This means that the gradient critic, which is most used at $\lambda \to 0$, helps in delivering high quality estimation of the gradient. Notably, the bias of semi-gradient affects only a subset of gradient vector (Figure~\ref{fig:imani-bias}c and d), suggesting that the gradient critic could be tailored to learn only a subset of the gradient (more details in Appendix~\ref{app:bias-variance}).\\
\textbf{Performance of LSTD$\bm{\Gamma}$.} The previous analysis supports the unbiasedness discussed in (Section~\ref{sec:unbiased}).  However, this does not automatically imply an increase in performance. To provide an analysis on the performance, we generated datasets of $500$ samples, and we trained the policy for $1000$ steps using the Adam optimizer \citep{kingma_adam_2014} with a learning rate of $0.01$. We repeat the process $20$ times for each value of $\lambda$.
Figure~\ref{fig:learning-curves}a depicts the final performance of the algorithm for the different values of $\lambda$. Observe that high values of $\lambda$, like $1$ or $0.9$, lead to a poor solution, while lower values of $\lambda$ reach high performance. This enhancement suggests that the contribution of the gradient critic is beneficial. Interestingly, even a weak mixing of the gradient critic helps the performance dramatically.\\
%
%
\textbf{Performance of TDRC$\bm{\Gamma}$.} The online estimator of the gradient function TDRC$\bm{\Gamma}$ requires a separate validation as it is more subject to noise in the data. We used a similar settings as for LSTD$\bm{\Gamma}$, except that samples are streamed, Adam's learning rate is set to $0.001$, and the optimization takes $5000$ steps. We used $\beta=1$ as regularization factor and constant learning rate for both critic and gradient critic $\alpha=0.1$. Figure~\ref{fig:learning-curves}b shows that the algorithm behaves similarly to LSTD$\bm{\Gamma}$, enforcing the idea that the gradient critic also helps when its approximation is more pronouced. The effect of the delayed gradient estimation does not impact negatively the performance (more details in Appendix~\ref{app:gamma-rc-imani}).
\subsection{Analysis on Randomly Generated MDPs}
Previous analyses show us a clear use-case where the gradient critic helps to solve the issue of semi-gradient approaches. Despite the convergence issue discussed by \citet{imani_off-policy_2018} and \citet{fujimoto_off-policy_2019}, semi-gradients are widely used since they perform reasonably well when the distribution shift is not too marked. Hence, we test TDRC$\bm{\Gamma}$ on $2500$ randomly generated MDPs.
We try to replicate realistic conditions and show that our method works well across different models. Our MDPs have 30 states and 2 actions. The structure of each MDP is generated randomly using a low-entropy distribution that ensures the sparsity of both mean reward and transitions. This sparsity ensures a diversification in the different MDPs, sometimes creating cycles and absorbing states.
We also add a Gaussian noise to the reward to make the setting more challenging. The discount factor is 0.95, while the episode length is 50 steps.

In this set of experiments, we do not provide a direct source of state aliasing. Instead, we codify each state with its numerical, 1-dimensional value. The actor, a neural network with one hidden layer of 5 neurons, receives complete information about the state. However, its under-parametrization (26 parameters in an MDP of 30 states and 2 actions per state) can cause a similar and more realistic aliasing effect (details in Appendix~\ref{app:tdrc-random-mdp}).

We test OffPAC, ACE(1), TDRC$\bm\Gamma$. In addition, we include TDRC$\bm\Gamma$-LastLayer, which uses the gradient critic only to update the last layer of the actor, while the remaining weights are updated with TDRC$\bm\Gamma$ with $\lambda$=1.
Figure~\ref{fig:random-mdps}c, shows that TDRC$\bm\Gamma$ outperforms both OffPAC and ACE(1) in this setting. Furthermore, lower values of $\lambda$ still obtain higher returns, showing that the gradient critic effectively improves the performance also in this scenario. It is interesting to notice that there is no substantial performance degradation between TDRC$\bm\Gamma$ and TDRC$\bm\Gamma$-LastLayer (Figure~\ref{fig:random-mdps}d), corroborating the intuition that applying the gradient critic only to the last layer of the actor is still beneficial. 
\section{Conclusion and Future Work}
Most policy gradient algorithms use off-policy samples without correcting the state distribution, causing a biased gradient estimate. Such bias deteriorates the algorithm's performance. 
Instead of resorting to importance sampling, we proposed to learn the policy gradient using a \textsl{gradient critic}. Like the classic value critic, our gradient critic is expressible with a Bellman equation, hence learnable via temporal-difference under off-policy distribution. The ability of the \textsl{gradient critic} to predict the gradient cumulation overcomes the need for sample reweighting. The gradient critic can provide an unbiased policy gradient estimator using arbitrary experience without resorting to importance sampling. Further, we introduced an approach based on eligibility traces that smoothly combines it with classic semi-gradient estimation. We showed empirically that our approach mitigates the high bias of semi-gradients, boosting its performance. 
Future work will focus on the extension of the gradient critic to deep reinforcement learning, using our technique to predict a subset of the policy gradient.
%
\pagebreak
\section*{Acknowledgment}

This research is financially supported by the Alberta Machine Intelligence Institute (Amii), the Canadian Institute for Advanced Research (CIFAR), the Reinforcement Learning and Artificial Laboratory (RLAI) in the university of Alberta, Canada.  We would like to thank Compute Canada for providing the computational resources needed.
\bibliography{zotero}
\bibliographystyle{icml2022}
\appendix
\onecolumn
\section{Supplement to the Theoretical Analysis}
This appendix is structured as follows: we introduce a \textsl{generalized} temporal-difference in Appendix~\ref{app:generalized-least-squares}, which will be useful to precisely determine Bellman equations and their least-squares solution. We prove Lemma~\ref{lemma:tdq} and Theorem~\ref{theo:GLS} in  Appendix~\ref{app:least-squares-gradient}. We prove Theorem~\ref{theo:perfect-features} in Appendix~\ref{app:perfect-features}.
\begin{algorithm}[t]
	\caption{Policy Gradient with LSTD$\bm{\Gamma}$ \label{alg:lstd-gamma}}
	\begin{algorithmic}[1]
		\STATE \textbf{Input:} Policy $\pi_\theta$, set of features $\phi$, learning rate $\eta$, and dataset $D$ of off-policy transitions $(\state_i, \action_i, r_i, \state_{i}')$.
		\STATE {$\hat{\mathbf{b}} =1/N \sum_i \phi(\state_i, \action_i) r_i$}
		\WHILE {not converged}
		\STATE For each $\state_i'$ sample $\action_i' \sim \pi_\theta(\cdot | \state_i')$
		\STATE {$\hat{\mathbf{A}} = 1/N \sum_i \phi(\state_i, \action_i)\left(\phi(\state_i, \action_i) - \gamma \phi(\state_i', \action_i')\right)^\intercal$}
		\STATE $\hat{Q}(\state, \action) = \phivec^\intercal(\state, \action)\hat{\omegavec}_{T\!D}$; $\hat{\omegavec}_{T\!D} = \hat{\mathbf{A}}^{-1} \hat{\mathbf{b}}$;
		\STATE {$\hat{\mathbf{B}} =1/N \sum_i \phi(\state_i, \action_i)\hat{Q}(\state_i', \action_i') \nabla_\theta\log\pi_\theta(\action_i'|\state_i')$}
		\STATE {$\hat{\bm{\Gamma}}(\state, \action) = \phivec^\intercal(\state, \action)\hat{\mathbf{G}}_{T\!D}$; $\hat{\mathbf{G}}_{T\!D} = \hat{\mathbf{A}}^{-1} \hat{\mathbf{B}}$}
		\STATE Sample $\state_0 \sim \mu_0$ (from dataset), $\action_0^\pi \sim \pi_\theta(\cdot|\state_0)$
		\STATE $\mathbf{g} = \hat{Q}(\state_0, \action_0^\pi) \nabla_\theta\log\pi_\theta(\action_0^\pi|\state_0)$
		\STATE $\theta \leftarrow\theta + \eta (\mathbf{g} + \hat{\bm{\Gamma}}(\state_0, \action_0^\pi ))$
		\ENDWHILE
	\end{algorithmic}
\end{algorithm}

\subsection{Gradient Function and Eligibility Traces}\label{app:eligibility-trace}
In this section, we detail all the passages to show the $n$-step view of the $\bm{\Gamma}$-function, and the eligibility-trace view. It is interesting to see the parallel between the $n$-step view and eligibility traces in critic estimation \cite{sutton_reinforcement_2018}.
\subsubsection{$n$-Step View of the Policy Gradient and Gradient Function}
\label{app:n-steps}
To start, let us remind that $\bm{\Gamma}^\pi(\state, \action) = \nabla_\theta Q^\pi(\state, \action)$ by definition.
We report here the \textsl{gradient} Bellman equation introduced in \eqref{eq:gbe}, and that can be seen as the application on both sides of \eqref{eq:bellman},
\begin{align}
	\bm{\Gamma}^\pi(\state, \action)=&\gamma\sum_{s' \in \Sset}\sum_{a' \in \Aset}\left( Q^\pi(\state', \action')\nabla_\theta\log\pi_\theta(\action'|\state') + \bm{\Gamma}^\pi(\state', \action')\right)\pi(\action'|\state')p(\state'|\state,\action). \nonumber
\end{align}
The LHS of the gradient Bellman equation can be expanded, by using the recursive definition of $\bm{\Gamma}(\state, \action)$,
\begin{align}
	\bm{\Gamma}^\pi(\state, \action)=&\gamma\sum_{s' \in \Sset}\sum_{a' \in \Aset}\Big( Q^\pi(\state', \action')\nabla_\theta\log\pi_\theta(\action'|\state') \nonumber\\
	& + \gamma\sum_{s'' \in \Sset}\sum_{a'' \in \Aset}\Big( Q^\pi(\state'', \action'')\nabla_\theta\log\pi_\theta(\action''|\state'') + \gamma \bm{\Gamma}^\pi(\state'', \action'')\Big)\pi(\action''|\state'')p(\state''|\state',\action')
	\Big)\pi(\action'|\state')p(\state'|\state,\action). \nonumber
\end{align}
Taking in consideration that $\sum_{s'' \in \Sset}\sum_{a'' \in \Aset} \pi(\action''|\state'')p(\state''|\state',\action') =1$, we can reformulate the gradient Bellman equation as
\begin{align}
	\bm{\Gamma}^\pi(\state, \action)=&\sum_{s', a', s'', a''}\Big( \gamma Q^\pi(\state', \action')\nabla_\theta\log\pi_\theta(\action'|\state') +  \gamma^2 Q^\pi(\state'', \action'')\nabla_\theta\log\pi_\theta(\action''|\state'') \nonumber\\
	& + \gamma^2 \bm{\Gamma}^\pi(\state'', \action'')\Big)
	\pi(\action''|\state'')p(\state''|\state',\action') \pi(\action'|\state')p(\state'|\state,\action), \nonumber
\end{align}
which is equivalent to
\begin{align}
	\bm{\Gamma}^\pi(\state, \action)=&\mathbb{E}_{\pi_\theta}\left[ \gamma Q^\pi(S_1, A_1)\nabla_\theta\log\pi_\theta(A_1|S_1) +  \gamma^2 Q^\pi(S_2, A_2)\nabla_\theta\log\pi_\theta(A_2|S_2) + \gamma^2 \bm{\Gamma}^\pi(S_2, A_2)\Big| S_0=\state, A_0=\action\right], \nonumber
\end{align}
This process can be repeated a finite number of time $n$, to find out that
\begin{align}
	\bm{\Gamma}^\pi(\state, \action)=&\mathbb{E}_{\pi_\theta}\left[ \sum_{t=1}^n \gamma^t Q^\pi(S_t, A_t)\nabla_\theta\log\pi_\theta(A_t|S_t) + \gamma^n \bm{\Gamma}^\pi(S_n, A_n)\Big| S_0=\state, A_0=\action\right]. \label{eq:n-step-gamma}
\end{align}
Equation~\ref{eq:first-derivation} united with Equation~\ref{eq:n-step-gamma} yields
\begin{align}
	\nabla_\theta J(\theta) = (1-\gamma) \underset{\tau_\pi}{\mathbb{E}}\bigg[\sum_{t=0}^{n-1} \gamma^{t}Q^\pi(S_t, A_t)\nabla_\theta \log \pi_\theta(A_t|S_t) + \gamma^{n-1}\bm{\Gamma}^\pi(S_{n-1}, A_{n-1})\bigg], \label{eq:n-step-gradient}
\end{align}
which is equivalent to the last passage in the derivation \ref{eq:gradient-critic-starting}.
\subsubsection{Eligibility-Trace View of Policy Gradient}
Consider $0 \leq \lambda < 1$. We know that $\sum_{n}^\infty \lambda^n = 1/(1-\gamma)$, hence $(1-\lambda) \sum_{n=0}^\infty \lambda^n = 1$. Consider not an enumerable set of expressions $\{x_n\}_i^\infty$ which are all mathematically equivalent to a value $x$, i.e., $x_0 = x_1 =  x_2 = \dots = x$.
We can say that $(1-\gamma)\sum_{n=0}^\infty \lambda^n x_n = x$.
Let
\begin{align}
	y_t := \mathbb{E}_{\tau_\pi}\left[Q^\pi(S_t, A_t)\nabla_\theta \log \pi_\theta(A_t | S_t)\right] \; \text{and} \; z_t := \mathbb{E}_{\tau_\pi} [\bm{\Gamma}(S_t, A_t){\color{violet}]}. \nonumber
\end{align}
In consideration of Equation~\ref{eq:n-step-gradient}, we can say that
\begin{align}
	\nabla_\theta J(\theta) =  (1-\gamma)(1-\lambda) \sum_{n=0}^\infty \lambda^n  \sum_{t=0}^{n} \gamma^t y_t + \gamma^n z_t.  \nonumber
\end{align}
The equation above can be rewritten by ``unrolling'' the innermost summation
\begin{align}
	\nabla_\theta J(\theta) = (1-\gamma)(1-\lambda) \bigg(& y_0 + z_0 \nonumber \\
	+ & \lambda y_0 + \lambda \gamma y_1 + \lambda \gamma z_1 \nonumber \\
	+ & \lambda^2 y_0 + \lambda^2 \gamma y_1 + \lambda^2 \gamma^2 y_2 + \lambda^2 \gamma^2 z_2 \nonumber \\
	+ &\lambda^3 y_0 + \lambda^3 \gamma y_1 + \lambda^3  \gamma^2 y_2 + \lambda^3 \gamma^3 y_3 + \lambda^3 \gamma^3 z_3   \nonumber \\
	+ &\lambda^4 y_0 + \lambda^4 \gamma y_1 + \lambda^4  \gamma^2 y_2 + \lambda^4 \gamma^3y_3 + \lambda^4 \gamma^4 y_4 + \lambda^4\gamma^4 z_4  + \dots \bigg) \nonumber
\end{align}
The equation has be graphically arranged to highlight its structure. In particular, we can see the right hand side as a summation of $y_n$ terms that can be collected together column-wise, plus a summation of $z_n$,
\begin{align}
	\nabla_\theta J(\theta) =  & (1-\gamma)\left( (1-\lambda)\sum_{n=0}^\infty \lambda^n y_0 + (1-\lambda)\sum_{n=1}^\infty \lambda^n \gamma y_1 + (1-\lambda)\sum_{n=2}^\infty \lambda^n \gamma^2 y_2 + \dots + (1-\lambda)\sum_{n=0}^\infty \lambda^n\gamma^n z_n \right)\nonumber \\
	=  & (1-\gamma)\left( (1-\lambda)\sum_{n=0}^\infty \lambda^n y_0 + (1-\lambda)\lambda \sum_{n=0}^\infty \lambda^n \gamma y_1 + (1-\lambda)\lambda^2 \sum_{n=0}^\infty \lambda^n \gamma^2 y_2 + \dots + (1-\lambda)\sum_{n=0}^\infty \lambda^n\gamma^n z_n\right) \nonumber \\
	=  & (1-\gamma)\left( (y_0 +\lambda \gamma y_1 + \lambda^2\gamma^2 y_2 + \dots ) (1-\lambda)\sum_{n=0}^\infty \lambda^n + (1-\lambda)\sum_{n=0}^\infty \lambda^n \gamma^nz_n\right) \nonumber \\
	=  & (1-\gamma)\left( (y_0 +\lambda \gamma y_1 + \lambda^2\gamma^2 y_2 + \dots )  + (1-\lambda)\sum_{n=0}^\infty \lambda^n \gamma^nz_n\right) \nonumber \\
	=  & (1-\gamma)\sum_{n=0}^\infty \lambda^n \gamma^n y_n  + (1-\lambda) \lambda^n \gamma^nz_n \nonumber \\
	=  & (1-\gamma)\sum_{n=0}^\infty \lambda^n \gamma^n\left( y_n  + (1-\lambda)z_n\right) \nonumber
\end{align}
Looking back at the definitions of $y_n$ and $z_n$, we can state
\begin{align}
	\nabla_\theta J(\theta) = &  (1-\gamma)\underset{\tau_\pi}{\mathbb{E}}\bigg[\sum_{n=0}^{\infty}\lambda^n\gamma^n\big(Q^\pi(S_n, A_n)\nabla_\theta \log \pi_\theta(A_n | S_n) + (1-\lambda) \bm{\Gamma}^\pi(S_n, A_n)\big)\bigg]. \nonumber
\end{align}
\subsection{Generalized Least-Squares Temporal-Difference}\label{app:generalized-least-squares}
This section provides a generalization of least-square temporal-difference. We introduce a setting that abstract the concepts of state and action (which will be seen as a conglomerate variable $x$), and that unifies a finite set of ``Bellman'' equations that share same dynamics but different ``rewards'' in a compact vectorial notation.
Eventually, we report the error analysis conduced by \cite{kolter_fixed_2011} using our vectorial notation.

\begin{proposition}[Generalized Least Squares]\label{prop:generalized-ls}
	Let us consider finite set $\{x_1, x_2, \dots, x_n\} \equiv \Xset$. Let us consider an irreducible Markov chain induced by the transition function $g(x'|x)$ with steady distribution $\mu$. Let us consider $K$ stochastic mappings $c_k:\Xset\to\Omega(\mathbb{R})$ where $\Omega(\mathbb{R})$ denotes the set of all probability distributions over $\mathbb{R}$. Let us assume that $\overline{c}_k(x) = \mathbb{E}[c(x)]$ exists and it is finite for all $x \in \Xset$ and $k \in \{1, \dots, K\}$. Consider $\gamma \in [0, 1)$. Consider the Bellman-like equations
	\begin{align}
		f_k(x) = \overline{c}(x) + \gamma  \sum_{x' \in \Xset} f_k(x')g(x'|x), \nonumber 
	\end{align}
	where each $f_k:\Xset \to \mathbb{R}$ exists and is unique. The equations above can be rewritten as
	\begin{align}
		\mathbf{f}(x) = \overline{\mathbf{c}}(x) + \gamma \sum_{x' \in \Xset} \mathbf{f}(x')g(x'|x), \label{eq:generalized-bellman}
	\end{align}
	where $\mathbf{f}:\Xset \to \mathbb{R}^K$ exists and is unique. Consider a function $\hat{\mathbf{f}}_t(x) = \phivec^\intercal(x)\mathbf{H}_t$ where $\phivec:\Xset\to\mathbb{R}^{n_f}$ is a feature vector and $\mathbf{H} \in \mathbb{R}^{n_f \times K}$. Furtermore, consider a matrix $\bm{\Phi}$ where each row $i$ is $\phivec^\intercal(x_i)$ and assume that all the columns of $\bm{\Phi}$ are linearly independend. Consider a process that starts with a desired parameter $\mathbf{H}_0$, and that updates
	\begin{align}
		\mathbf{H}_{t+1}\! =\! \argmin_{\mathbf{H}}\! \|\phivec^\intercal(x)\mathbf{H} \! -\! \overline{\mathbf{c}}(x)\! -\! \gamma \! \sum_{x' \in \Xset}\!\hat{\mathbf{f}}_t(x')g(x'|x)\|_{d}, \label{eq:vectorial-projection}
	\end{align}
	where $\|\mathbf{x}\|_{d} = \mathbb{E}_d[\langle\mathbf{x}, \mathbf{x}\rangle]$.
	It is possible to verify that the process described in \eqref{eq:vectorial-projection} is equivalent to
	\begin{align}
		\mathbf{h}_{t+1, i}\! =\! \argmin_{\mathbf{h}}\! \|\phivec^\intercal(x)\mathbf{h} \! -\! \overline{c}_i(x)\! -\! \gamma \! \sum_{x' \in \Xset}\!\hat{f}_{t, i}(x')g(x'|x)\|_{d} \label{eq:scalar-projection}
	\end{align}
	with $\mathbf{H}_{t+1} = [\mathbf{h}_{t+1, 1}, \mathbf{h}_{t+1, 2}, \dots, \mathbf{h}_{t+1, k}]$.
	As reported by \cite{lagoudakis_least-squares_2003}, the fixed point of \eqref{eq:scalar-projection} is
	\begin{align}
		\mathbf{h}^*_i  \!=\! \underset{\substack{x \sim d,\\
				x' \sim g(x)}}{ \mathbb{E}}\left[\phivec(x)\!\left(\phivec(x) \!-\! \gamma\phivec(x')\right)^\intercal\right]^{-1}\!\mathbb{E}_{x \sim d}\left[\phivec(x)\overline{c}_i(x)\right]\nonumber
	\end{align},
	which can be compactly rewritten in vectorial notation
	\begin{align}
		\mathbf{H}^* \!=\! \underset{\substack{x \sim d,\\
				x' \sim g(x)}}{ \mathbb{E}}\left[\phivec(x)\!\left(\phivec(x) \!-\! \gamma\phivec(x')\right)^\intercal\right]^{-1}\!\mathbb{E}_{x \sim d}\left[\phivec(x)\mathbf{c}^\intercal(x)\right].\label{eq:generalized-least-squares}
	\end{align}
	Thanks to the work of \cite{kolter_fixed_2011}, we are able to bound the ``scalar'' fixed point solution, i.e.,
	\begin{align}
		\|\phivec^\intercal(x)\mathbf{h}^* - f_i(x)\|_{d} \leq \frac{1 + \kappa\gamma }{1-\gamma}\min_{\mathbf{h}} \|\phivec^\intercal(x)\mathbf{h} - f_i(x)\|_d.\nonumber
	\end{align}
	where $\kappa = \max_i \sqrt{d(x_i)/\mu(x_i)}/ \min_i \sqrt{d(x_i)/\mu(x_i)}$ and $d$ satisties the inequality in \cite{kolter_fixed_2011}.
	Knowing that $\|\xvec\|_d = \mathbb{E}_d[\langle \xvec, \xvec\rangle] = \sum_i \mathbb{E}_d[x_i^2]$, we can see that
	\begin{align}
		& \sum_i \|\phivec^\intercal(x)\mathbf{h}^*_i - f_i(x)\|_{d} \leq \sum_i\frac{1 + \kappa\gamma }{1-\gamma}\min_{\mathbf{h}} \|\phivec^\intercal(x)\mathbf{h} - f_i(x)\|_d \nonumber  \\
		\implies & \|\phivec^\intercal(x)\mathbf{H}^* - \mathbf{f}(x)\|_{d} \leq \frac{1 + \kappa\gamma }{1-\gamma}\min_{\mathbf{H}} \|\phivec^\intercal(x)\mathbf{H} - \mathbf{f}(x)\|_d. \label{eq:kolter-vectorial}
	\end{align}
\end{proposition}
\subsection{Least Squares Solution for the Gradient Function}\label{app:least-squares-gradient}
\begin{proof}[Proof of Lemma~\ref{lemma:tdq}]
	Let us analize Equation~\ref{eq:gbe} for a single parameter $\theta_k$ and for finite state-action space,
	\begin{align}
		\Gamma^\pi_k(\state, \action)=\gamma\sum_{\state'}\sum_{\action'}\Big( Q^\pi(\state', \action')\frac{\partial}{\partial \theta_k}\log\pi_\theta(\action'|\state') +  \Gamma^\pi_k(\state', \action')\Big)\pi(\action'|\state')p(\state'|\state,\action)
	\end{align}
	Let us set $\xvec = (\state, \action)$ and $g(\xvec' | \xvec) = \pi(\action'| \state')p(\state'|\state, \action)$, $e(\xvec) = Q(\state, \action)\partial/\partial\theta_k\pi_\theta(\action|\state)$, and $\overline{c}(\xvec) = \sum_i e_i(\xvec')g(\xvec'|\xvec)$, and, by posing $f(\xvec) = \Gamma_i^\pi(\state, \action)$, we realize that
	\begin{align}
		&	\Gamma^\pi_k(\state, \action)=\gamma\sum_{\state'}\sum_{\action'}\Big( Q^\pi(\state', \action')\frac{\partial}{\partial \theta_k}\log\pi_\theta(\action'|\state') +  \Gamma^\pi_k(\state', \action')\Big)\pi(\action'|\state')p(\state'|\state,\action)\nonumber \\
		\implies  &f(\xvec) = \overline{c}(\xvec) + \gamma f(\xvec_i) p(\xvec_i|\xvec).
	\end{align}
	Notice that in Lemma~\ref{lemma:tdq}, we consider the stationary distribution $\mu_\pi$ w.r.t. the transition $p(s'| s, a)$. we notice that $\mu_\pi(\state)\pi(\action|\state)$ is the stationary distribution w.r.t. thetransition $g(\xvec' | \xvec)$. Taking in consideration Equation~\ref{eq:generalized-least-squares}, we can prove that $\hat{\bm{\Gamma}}^\pi_{TDQ}$ is a fixed point of the approximated gradient Bellman equation. Furthermore, it is possible to prove, thank to \cite{kolter_fixed_2011}, that
	\begin{align}
		& \left\| \bm{\Gamma}^\pi_{T\!D\!Q, k}(\state, \action) - \frac{\partial}{\partial \theta_k} Q^\pi(\state, \action)\right\|_{\zeta} \leq \frac{1 - \gamma\kappa}{1-\gamma} \min_{\mathbf{g}}\left\|\phivec^\intercal(\state,\action)\mathbf{g} - \frac{\partial}{\partial \theta_k} Q^\pi(\state, \action)\right\|_{\zeta}, \nonumber
	\end{align}
	and, therefore,
	\begin{align}
		& \left\| \bm{\Gamma}^\pi_{T\!D\!Q}(\state, \action) - \nabla_\theta Q^\pi(\state, \action)\right\|_{\zeta} \leq \frac{1 - \gamma\kappa}{1-\gamma} \min_{\mathbf{G}}\left\|\phivec^\intercal(\state,\action)\mathbf{G} - \nabla_\theta Q^\pi(\state, \action)\right\|_{\zeta}.
	\end{align}
	where $\kappa\! =\! \max_{\state, \action}h(\state, \action)/\min_{\state,\action} h(\state, \action)$, $ h(\state, \action) = \sqrt{\mu(\state)\pi_\theta(\action|\state)}/\sqrt{\mu_\beta(\state)\beta(\action|\state))}$ and $\mu_\pi(\state)\pi_\theta(\action | \state)$ must comply the matrix inequality defined in \cite{kolter_fixed_2011}.
\end{proof}
\begin{proof}[Proof of Theorem~\ref{theo:GLS}]
	Let us consider an arbitrary order of state and action pairs, and a feature matrix
	\begin{align}
		\bm{\Phi} = \begin{bmatrix}
			\phivec^\intercal(\state_1, \action_1) \\
			\phivec^\intercal(\state_1, \action_2) \\
			\vdots \\
			\phivec^\intercal(\state_n, \action_m)
		\end{bmatrix},
		\bm{\varPhi} = \begin{bmatrix}
			\bm{\varphi}^\intercal(\state_1, \action_1) \\
			\bm{\varphi}^\intercal(\state_1, \action_2) \\
			\vdots \\
			\bm{\varphi}^\intercal(\state_n, \action_m)
		\end{bmatrix} \nonumber
	\end{align}
	where $n$ is the number of states and $m$ is the number of actions. $\bm{\Phi} \in \mathbb{R}^{nm \times n_f}$.
	Consider an arbitrary parameter vector $\omegavec$. $\bm\varPhi\omegavec$ returns a vector of values for each state-action pairs.
	We denote the TD solution of the $Q$-function with $\hat{\mathbf{q}} = \bm{\varPhi}\omegavec_{T\!D}$. Pairwise, we denote a matrix representing the matrix function with $\hat{\bm{\nu}} = \bm{\Phi}\mathbf{G}_{T\!D}$. The true $Q$-function and $\bm{\Gamma}$-function are, in vector notation,
	\begin{align}
		\mathbf{q}^\pi = \begin{bmatrix}
			Q^\pi(\state_1, \action_1) \\
			Q^\pi(\state_1, \action_2) \\
			\vdots \\
			Q^\pi(\state_n, \action_m)
		\end{bmatrix},  \nonumber \quad 	\bm{\nu}^\pi = \begin{bmatrix}
			\bm{\Gamma}^\pi(\state_1, \action_1) \\
			\bm{\Gamma}^\pi(\state_1, \action_2) \\
			\vdots \\
			\bm{\Gamma}^\pi(\state_n, \action_m)
		\end{bmatrix}.
	\end{align}
	Similarly to \cite{lagoudakis_least-squares_2003}, we introduce  the transition matrix $\mathbf{P}$ and the policy $\bm{\Pi}$
	\begin{align}
		\bm{\Pi} = \mathbf{I}_n \otimes \mathbf{\pi}^\intercal \quad \text{where} \quad \bm{\pi} = [\pi_\theta(\action_1 |\state_1), \pi_\theta(\action_2 | \state_1), \dots, \pi_\theta(\action_m | \state_n)]^\intercal\nonumber
	\end{align}
	and
	\begin{align}
		\mathbf{P} = \begin{bmatrix}
			\mathbf{P}_1 \\
			\mathbf{P}_2 \\
			\vdots \\
			\mathbf{P}_n
		\end{bmatrix}\quad \text{where}\quad  \mathbf{P}_i =  \begin{bmatrix}
			p(\state_i |\state_1, \action_1) & p(\state_i |\state_2, \action_1) & \dots &   p(\state_i |\state_n, \action_1)\\
			p(\state_i |\state_1, \action_2) & p(\state_i |\state_2, \action_2) & \dots &   p(\state_i |\state_n, \action_2)\\
			\multicolumn{3}{c}{\vdots} \\
			p(\state_i |\state_1, \action_m) & p(\state_i |\state_2, \action_m) & \dots &   p(\state_i |\state_n, \action_m)
		\end{bmatrix}
	\end{align}
	Let $D$ a diagonal matrix where at each entry we have $\mu_\beta(x_i)\beta(a_j| x_i)$ where the indexes follow the enumeration introduced above. Let us introduce the norm  $\| \mathbf{M} \|_{D} $ of a matrix $\mathbf{M}$,
	\begin{align}
		\| \mathbf{M} \|_{D} = \sqrt{\sum_i D_{i, i} \langle \mathbf{M}_i, \mathbf{M}_i \rangle } \nonumber
	\end{align}
	The least squares solution of the gradient inder the norm $\|\cdot\|_D$ is the unique solution of
	\begin{align}
		\hat{\bm{\nu}} = \bm{\Psi} \left(\gamma \bm{\Pi}\mathbf{P}(\nabla_\theta\log\bm{\pi})\odot\hat{\mathbf{q}} + \gamma \bm{\Pi}\mathbf{P}\hat{\bm{\nu}} \right)
	\end{align}
	where
	\begin{align}
		\nabla_\theta\log\bm{\pi} = \begin{bmatrix}
			\nabla_\theta^\intercal\log\pi_\theta(\action_1 | \state_1) \\
			\nabla_\theta^\intercal\log\pi_\theta(\action_2 | \state_1) \\
			\vdots \\
			\nabla_\theta^\intercal\log\pi_\theta(\action_m | \state_n)
		\end{bmatrix}, \nonumber
	\end{align}
	$(\mathbf{A}  \odot \mathbf{b})_i = \mathbf{A}_i b_i$ is a row-wise product and $\bm{\Psi}_D=\bm{\Phi}(\bm{\Phi}^\intercal D \bm{\Phi})^{-1}\bm{\Phi}^\intercal D$ is a non-expansion under the norm $\|\cdot\|_D$ as shown by \cite{tsitsiklis_analysis_1997}. Remember, that $\bm{\Psi}$ is a least-square projection under the norm $\|\cdot\|_D$, and, therefore, $\|\bm{\Psi}\mathbf{M} - \mathbf{M}\|_D = \min_\mathbf{H}\|\bm{\Phi}\mathbf{H} - \mathbf{M}\|$.
	The true gradient function is the fixed point of the gradient Bellman equation,
	\begin{align}
		\bm{\nu}^\pi = \gamma \bm{\Pi}\mathbf{P}(\nabla_\theta\log\bm{\pi})\odot\mathbf{q}^\pi + \gamma \bm{\Pi}\mathbf{P} \bm{\nu}.
	\end{align}
	We want now to bound $\|\hat{\bm{\nu}} - \bm{\nu}^\pi\|_D$.
	\begin{align}
		\|\hat{\bm{\nu}} - \bm{\nu}^\pi\|_D & = \|\hat{\bm{\nu}} - \bm{\Psi}\bm{\nu}^\pi + \bm{\Psi}\bm{\nu}^\pi  - \bm{\nu}^\pi\|_D \nonumber \\
		&\leq  \|\hat{\bm{\nu}} - \bm{\Psi}\bm{\nu}^\pi\|_D +\| \bm{\Psi}\bm{\nu}^\pi  - \bm{\nu}^\pi\|_D \nonumber\\
		&=  \|\bm{\Psi}\hat{\bm{\nu}} - \bm{\Psi}\bm{\nu}^\pi\|_D +\| \bm{\Psi}\bm{\nu}^\pi  - \bm{\nu}^\pi\|_D \nonumber\\
		&=  \|\bm{\Psi}\left(\gamma \bm{\Pi}\mathbf{P}(\nabla_\theta\log\bm{\pi})\odot\hat{\mathbf{q}} + \gamma \mathbf{P}\Pi \hat{\bm{\nu}} \right) - \bm{\Psi}\left(\gamma \bm{\Pi}\mathbf{P}(\nabla_\theta\log\bm{\pi})\odot\hat{\mathbf{q}} + \gamma \mathbf{P}\Pi \bm{\nu}^\pi \right)\|_D +\| \bm{\Psi}\bm{\nu}^\pi  - \bm{\nu}^\pi\|_D \nonumber \\
		&\leq \gamma\underbrace{ \| \bm{\Psi}\mathbf{P}\Pi (\hat{\bm{\nu}} - \bm{\nu}^\pi) \|_D}_\text{A} + \gamma \underbrace{\|\bm{\Psi}\bm{\Pi}\mathbf{P}(\nabla_\theta\log\bm{\pi})\odot(\hat{\mathbf{q}} - \mathbf{q}^\pi)\|_D}_\text{B}  + \| \bm{\Psi}\bm{\nu}^\pi  - \bm{\nu}^\pi\|_D \nonumber\nonumber
	\end{align}
	\begin{table}
		\centering
		\caption{Description of symbols used in proof of Theorem~\ref{theo:GLS}.}
		\begin{tabular}{|l|c|l|}
			\hline
			Symbol & Dimension & Meaning \\
			\hline
			$n$ & - & Number of states \\
			$m$ & - & Number of actions \\
			$n_p$ & - & Number of policy parameters \\
			$n_p$ & - & Number of features \\
			$\gamma$ & - & Discount factor \\
			b & - &  $| \max_{\action, \state, i} \partial/\partial\theta_i\log\pi(\action|\state)|$ \\
			$\kappa$ & - & Defined in Lemma~\ref{lemma:tdq} \\
			D  & $nm \times nm$ & Diagonal matrix containing off-policy probabilities $\mu_\beta(s_i)\beta(a_j|s_i )$\\
			$\mathrm{q}$ & $nm \times 1$ & Vector of $Q$-values \\
			$\hat{\mathrm{q}}$ & $nm \times 1$ & TD-solution of $\mathrm{q}$ \\
			$\bm{\Phi}$ & $nm \times n_f$ & Matrix of features \\
			$\bm{\Psi}$ & $nm \times nm$ & Orthogonal projection onto $\|\cdot\|_D$\\
			$\mathbf{P}$ & $n \times n m $ & Transition matrix \\
			$\bm{\pi}$ & $nm \times 1$ & Vector represention of the policy \\
			$\bm{\Pi}$ &$nm \times n$ & Matrix representation of the olicy \\
			$\nabla_{\theta}\log\bm{\pi}$ & $nm \times n_p$ & Matrix of gradients of $\log \bm{\pi}$ \\
			$\bm{\nu}$ & $nm \times n_p$ & Matrix representing the true $\bm{\Gamma}$ per state-action pairs \\
			$\hat{\bm{\nu}}$ & $nm \times n_p$ & TD-solution of $\bm{\Gamma}$ per state-action pairs \\
			\hline
		\end{tabular}
	\end{table}
	\textbf{Upperbound of term A.} Since $D$ satisfies, by assumption, the inequality in \citep{kolter_fixed_2011}, then $\| \bm{\Psi}\bm{\Pi}\mathbf{P}\bm{\Phi}\omegavec\||_D \leq \|\bm{\phi}\omegavec\|_D$,
	\begin{align}
		\| \bm{\Psi}\mathbf{P}\bm{\Pi} (\hat{\bm{\nu}} - \bm{\nu}^\pi) \|_D = \| \bm{\Psi}\bm{\Pi}\mathbf{P} (\bm{\Phi}\mathbf{G}_{TD} - \bm{\nu}^\pi) \|_D \leq \| \bm{\Psi}\bm{\Pi}\mathbf{P}( \bm{\Phi}\mathbf{G}_{TD} - \bm{\Psi}\bm{\nu}^\pi)\|_D  + \|\bm{\Psi}\bm{\Pi}\mathbf{P}( \bm{\Psi}\bm{\nu}^\pi- \bm{\nu}^\pi) \|_D
	\end{align}
	knowing that exists some $\overline{\mathbf{G}}$ such that $ \bm{\Psi}\bm{\nu}^\pi = \bm{\Phi}\overline{\mathbf{G}}$, we have that
	\begin{align}
		& \| \bm{\Psi}\bm{\Pi}\mathbf{P}( \bm{\Phi}\mathbf{G}_{TD} - \bm{\Psi}\bm{\nu}^\pi)\|_D  + \|\bm{\Psi}\bm{\Pi}\mathbf{P}( \bm{\Psi}\bm{\nu}^\pi- \bm{\nu}^\pi) \|_D   \nonumber \\
		= & \| \bm{\Psi}\bm{\Pi}\mathbf{P}\bm{\Phi}(\mathbf{G}_{TD} - \overline{\mathbf{G}})\|_D  + \|\bm{\Psi}\bm{\Pi}\mathbf{P}( \bm{\Psi}\bm{\nu}^\pi- \bm{\nu}^\pi) \|_D  \nonumber \\
		\leq & \|\bm{\Phi}\mathbf{G}_{TD} - \bm{\Phi}\overline{\mathbf{G}}\|_D  + \|\bm{\Psi}\bm{\Pi}\mathbf{P}( \bm{\Psi}\bm{\nu}^\pi- \bm{\nu}^\pi) \|_D  \nonumber
	\end{align}
	furthermore, thanks to the convexity of the spanning set, we have that
	\begin{align}
		\|\bm{\Phi}\mathbf{G}_{TD} - \bm{\Phi}\overline{\mathbf{G}}\|_D  \leq  \|\hat{\bm{\nu}}- \bm{\nu}^\pi\|_D. \nonumber
	\end{align}
	Furthermore,
	\begin{align}
		\|\bm{\Psi}\bm{\Pi}\mathbf{P}( \bm{\Psi}\bm{\nu}^\pi- \bm{\nu}^\pi) \|_D \leq 	\|\bm{\Pi}\mathbf{P}\|_D \| \bm{\Psi}\bm{\nu}^\pi- \bm{\nu}^\pi \|_D, \nonumber
	\end{align}
	which yields
	\begin{align}
		\| \bm{\Psi}\bm{\Pi}\mathbf{P} (\hat{\bm{\nu}} - \bm{\nu}^\pi) \|_D \leq \|\hat{\bm{\nu}} - \bm{\nu}^\pi\|_D +   \|\bm{\Pi}\mathbf{P}\|_D \| \bm{\Psi}\bm{\nu}^\pi- \bm{\nu}^\pi \|_D
	\end{align}
	\textbf{Upperbound of term B.} The upperbound of the term B follows a very similar structure to term A, with the addition that we need to deal with the weighting $\nabla_\theta \log \bm{\pi}$. Recalling the non-expansion from \cite{kolter_fixed_2011} and the definition of the row-wise product $\odot$  given earlier,
	\begin{align}
		\| \bm{\Psi}\bm{\Pi}\mathbf{P} (\nabla_\theta\log\bm{\pi})\odot(\hat{\mathbf{q}} - \mathbf{q}^\pi)\|_D \leq
		& \| \bm{\Pi}\mathbf{P}\|_D \| (\nabla_\theta\log\bm{\pi})\odot(\hat{\mathbf{q}} - \mathbf{q}^\pi) \|_D \nonumber \\
		\leq & n_{p} \|\nabla_\theta \log\bm{\pi}\|_{\infty} \| \bm{\Pi}\mathbf{P}\|_D \| (\hat{\mathbf{q}} - \mathbf{q}^\pi) \|_D,\nonumber
	\end{align}
	where $n_d$ is the number of parameters of the policy.
	The quantity $\| \hat{\mathbf{q}} - \mathbf{q}^\pi \|_D$ has been bounded in Lemma~\ref{lemma:tdq},
	\begin{align}
		\| \bm{\Psi}\bm{\Pi}\mathbf{P} (\nabla_\theta\log\bm{\pi})\odot(\hat{\mathbf{q}} - \mathbf{q}^\pi)\|_D	\leq & \|\nabla_\theta \log\bm{\pi}\|_{\infty} \| \bm{\Pi}\mathbf{P}\|_D \frac{1 +\gamma\kappa}{1-\gamma} \min_{\omegavec}\|\bm{\varPhi}\omegavec - \mathbf{q}^\pi\| .\nonumber
	\end{align}
	we notice that the term $b$ introduced in Theorem~\ref{theo:GLS} is actually $\|\nabla_\theta \log\bm{\pi}\|_{\infty}$, hence,
	\begin{align}
		\| \bm{\Psi}\bm{\Pi}\mathbf{P} (\nabla_\theta\log\bm{\pi})\odot(\hat{\mathbf{q}} - \mathbf{q}^\pi)\|_D	\leq & b \| \bm{\Pi}\mathbf{P}\|_D \frac{1 +\gamma\kappa}{1-\gamma} \min_{\omegavec}\|\bm{\varPhi}\omegavec - \mathbf{q}^\pi\| .\nonumber
	\end{align}
	\textbf{Collecting both upperbound of terms A and B.}
	\begin{align}
		\|\hat{\bm{\nu}} - \bm{\nu}^\pi\|_D \leq & \gamma \|\hat{\bm{\nu}}- \bm{\nu}^\pi\|_D  + \gamma \underbrace{\|\bm{\Psi}\bm{\Pi}\mathbf{P}(\nabla_\theta\log\bm{\pi})\odot(\hat{\mathbf{q}} - \mathbf{q}^\pi)\|_D}_\text{B}  + (1 + \gamma \|\bm{\Pi}\textbf{P}\|_D)\| \bm{\Psi}\bm{\nu}^\pi  - \bm{\nu}^\pi\|_D \nonumber\nonumber
	\end{align}
	Notice that the term $\|\bm{\Pi}\mathbf{P}\|_D$ can be bounded by $\kappa$, as illustrated in \cite{kolter_fixed_2011}, and therefore
	\begin{align}
		\|\hat{\bm{\nu}} - \bm{\nu}^\pi\|_D \leq & \gamma \|\hat{\bm{\nu}}- \bm{\nu}^\pi\|_D  + \gamma \underbrace{\|\bm{\Psi}\bm{\Pi}\mathbf{P}(\nabla_\theta\log\bm{\pi})\odot(\hat{\mathbf{q}} - \mathbf{q}^\pi)\|_D}_\text{B}  + (1 + \gamma \kappa ) \| \bm{\Psi}\bm{\nu}^\pi  - \bm{\nu}^\pi\|_D. \nonumber
	\end{align}
	Projection errors like $\|\bm{\Psi}\mathbf{q}^\pi - \mathbf{q}^\pi\|_D$ and $\|\bm{\Psi}\bm{\nu}^\pi - \bm{\nu}^\pi\|_D$ can be bounded by
	\begin{align}
		\|\bm{\Psi}\mathbf{q}^\pi - \mathbf{q}^\pi\|_D \leq \min_{\omegavec} \| \bm{\varPhi}\omegavec - \mathbf{q}^\pi\|_D \quad \text{and} \quad \|\bm{\Psi}\bm{\nu}^\pi - \bm{\nu}^\pi\|_D \leq \min_{\mathbf{G}} \|\bm{\Phi}\mathbf{G}- \bm{\nu}^\pi\|_D. \nonumber
	\end{align}
	Hence,
	\begin{align}
		(1-\gamma)\|\hat{\bm{\nu}} - \bm{\nu}^\pi\|_D \leq &  \gamma \underbrace{\|\bm{\Psi}\bm{\Pi}\mathbf{P}(\nabla_\theta\log\bm{\pi})\odot(\hat{\mathbf{q}} - \mathbf{q}^\pi)\|_D}_\text{B}  + (1 + \gamma \kappa )  \min_{\mathbf{G}} \|\bm{\Phi}\mathbf{G}- \bm{\nu}^\pi\|_D. \nonumber \\
		\leq& \gamma n_p  b \kappa  \frac{1 +\gamma\kappa}{1-\gamma} \min_{\omegavec}\|\bm{\varPhi}\omegavec - \mathbf{q}^\pi\| + (1 + \gamma \kappa )  \min_{\mathbf{G}} \|\bm{\Phi}\mathbf{G}- \bm{\nu}^\pi\|_D. \nonumber
	\end{align}
	which yields
	\begin{align}
		\|\hat{\bm{\nu}} - \bm{\nu}^\pi\|_D
		\leq&   \gamma  n_p b\kappa\frac{1 + \gamma\kappa}{(1-\gamma)^2} \min_{\omegavec}\|\bm{\varPhi}\omegavec - \mathbf{q}^\pi\|  + \frac{(1 + \gamma \kappa )}{1-\gamma}  \min_{\mathbf{G}} \|\bm{\Phi}\mathbf{G}- \bm{\nu}^\pi\|_D.\nonumber
	\end{align}
\end{proof}
\subsection{Unbiased Gradient with Perfect Features}\label{app:perfect-features}
\begin{proof}[Proof of \textbf{Theorem~\ref{theo:perfect-features}}]
	We start the proof by showing that
	\begin{align}
		\phivec^\intercal(\state, \action) \omega_{TD} = Q^\pi(\state, \action) \nonumber
	\end{align}
	Notice that by assumption, there must be a vector $\omegavec$ such that
	\begin{align}
		\xivec^\intercal(\state, \action)\omegavec = r(\state, \action). \label{eq:proof:linear-space}
	\end{align}
	where
	\begin{align}
		\xivec(\state, \action) = \phivec(\state, \action) -\gamma \sum_{s \in \Sset} \sum_{a \in \Aset} \phivec(\state', \action')\pi_\theta(\action'|\state')p(\state'|\state, \action)\de\action' \de\state'
	\end{align}
	Since $\Phi'$ is a $n_f$-dimensional vector space, there is one and only one $\omega$ satisfating the relation above. Given the fact that \eqref{eq:proof:linear-space} is a linear equation, we can take a set of linearly independent features $\xivec$ to solve it.
	Since $\Phi'$ admits a $n_f$-dimensional basis, there exist $\{(s_i, s_i)\}_{i=1}^{n_f}$ such that we can construct a set of $n_d$ linearly independent vectors $\mathbf{e}_i = \xivec(\state_i, \action_i)$.
	Let us construct a basis matrix $\mathbf{E} = [\mathbf{e}_1, \mathbf{e}_2, \dots, \mathbf{e}_{n_f}]$.
	The unique solution of \eqref{eq:proof:linear-space} is determined by
	\begin{align}
		\omegavec^* = \mathbf{E}^{-\intercal}\mathbf{r}, \label{eq:proof:otd-sol}
	\end{align}
	where $\mathbf{r} = [r(\state_1, \action_1), r(\state_2, \action_2), \dots, r(\state_{n_f}, \action_{n_f})]^\intercal$.
	The TD solution satisfies
	\begin{align}	&\mathbb{E}_\zeta\left[\phivec(S, A)\left(\phivec^\intercal(S ,A) - \gamma  \phivec^\intercal(S', A') \right)\right]\omegavec_{TD}  = \mathbb{E}_\zeta\left[\phivec(S, A)r(S, A)\right] \nonumber \\
		\implies & \mathbb{E}_\zeta\left[\phivec(S, A)\xivec^\intercal(S, A)\right]\omegavec_{TD}  = \mathbb{E}_\zeta\left[\phivec(S, A)r(S, A)\right], \label{eq:proof:otd-1}
	\end{align}
	where $\zeta$ is a process generating $S \sim \mu_\beta, A \sim \beta(\cdot|S), S' \sim p(\cdot | S, A)$ and $A' \sim \pi_\theta(\cdot |S')$.
	Notice that, thanks to the property of vector spaces, there are two functions $\mathbf{f}:\Sset\times\Aset \to \mathbb{R}^{n_f}$ and $\mathbf{h}:\Sset\times\Aset \to \mathbb{R}^{n_f}$ such that
	\begin{align}
		& \xivec(\state, \action) = \mathbf{E} \mathbf{f}(\state, \action),  \phivec(\state, \action) = \mathbf{B} \mathbf{h}(\state, \action) \tag*{$\forall \state \in \Sset \land \action \in \Aset$,}
	\end{align}
	where $\mathbf{B}$ is a basis function for the set $\Phi$ (defined in Theorem~\ref{theo:perfect-features}).
	We can rewrite \eqref{eq:proof:otd-1} as
	\begin{align}
		\mathbb{E}_\zeta\left[\mathbf{B}\mathbf{h}(S, A)\mathbf{f}^\intercal(S, A)\mathbf{E}^\intercal \right]\omegavec_{TD}  = \mathbb{E}_\zeta\left[\mathbf{B}\mathbf{h}(S, A)r(S, A)\right]  \nonumber
	\end{align}
	looging back to Equation~\ref{eq:proof:linear-space}, we notice that $r(S, A) = f^\intercal(S, A)\mathbf{E}^\intercal \omega = f^\intercal(S, A)\mathbf{E}^\intercal \mathbf{E}^{-\intercal}\mathbf{r}$, and, therefore,
	
	\begin{align}
		& \mathbb{E}_\zeta\left[\mathbf{B}\mathbf{h}(S, A)\mathbf{f}^\intercal(S, A)\mathbf{E}^\intercal \right]\omegavec_{TD}  = \mathbb{E}_\zeta \left[\mathbf{B}\mathbf{h}(S, A)f^\intercal(S, A)\mathbf{E}^\intercal \mathbf{E}^{-1}\mathbf{r}\right]  \nonumber \\
		\implies & \omegavec_{TD}  = \mathbf{E}^{-\intercal}\mathbf{r}.  \nonumber
	\end{align}
	Therefore, looking back at Equation~\ref{eq:proof:otd-sol}
	\begin{align}
		\phivec^\intercal(\state, \action)\omegavec_{TD} = \phivec^\intercal(\state, \action)\omegavec^* =   Q^\pi(\state, \action) \tag*{$\forall \state\in\Sset \land \action \in \Aset$.}
	\end{align}
	This result is valid for any policy $\pi$, and state-action pairs. This implies that
	\begin{align}
		\nabla_\theta \phivec^\intercal(\state, \action)\omegavec_{TD}= \nabla_\theta Q^\pi(\state, \action) \tag*{$\forall \state\in\Sset \land \action \in \Aset$,}
	\end{align}
	which, thanks to Lemma~\ref{lemma:gradient-td}, implies that
	\begin{align}
		\bm{\Gamma}(\state, \action) = \nabla_\theta Q^\pi(\state, \action) \tag*{$\forall \state\in\Sset \land \action \in \Aset$.}
	\end{align}
\end{proof}
\newpage
\section{Extension to the Continuous State-Action Space}\label{app:continuous-action}
Consider a Markov decision process formed by the tuple $(\Sset, \Aset, r, p, \gamma, \mu_0)$
where  $\Sset$ and $\Aset$ represent the set of states and actions, $r:\Sset\times\Aset\to [-R_{\max}, R_{\max}]$ is a bounded reward function, $p:\Sset\times\Aset \to \mathcal{M}(\Sset)$ is a transition probability, $\gamma \in [0, 1)$ the discount factor and $\mu_0 \in \mathcal{M}(\Sset)$ a distribution of starting states. We assume that the policy $\pi_\theta:\Sset \to \mathcal{M}(\Aset)$ is differentiable w.r.t. its parameters $\theta$. We denoted  with $\mathcal{M}(X)$ the set of probability measures over a $\sigma$-algebra on a set $X$.

\textbf{Informal Extension of the Proofs to Continuous State-Action Space.} The proofs in Appendix~\ref{app:n-steps},~\ref{app:perfect-features} remain valid in the continuous case, since they only require substituting summations with integrals.
The proofs in Appendix~\ref{app:generalized-least-squares}~\ref{app:least-squares-gradient} can also be arranged in the continuous state-action spaces by rewriting the norm operator $\|\mathrm{B}\|_d = \sqrt{d_i \langle\mathbf{b}_i, \mathbf{b}_i\rangle}$ as $\sqrt{\mathbb{E}_{\bvec \sim d(\bvec)}\left[\langle\mathbf{b}, \mathbf{b}\rangle\right]}$ with $\mathbf{b} \in \mathcal{B}$ where $d$ is a probability measure over a $\sigma$-algebra on $\mathcal{B}$. 

\subsection{Reparametrization Gradient}
The gradient Bellman equation can be framed also in terms of reparametrization gradient. Suppose that we have a function
$f(\state, \epsilon)$ with $\epsilon \sim p$ such that
\begin{align}
	A = f_\theta(\state, \epsilon) \ {\buildrel d \over =} \ A \sim \pi_\theta(\cdot | \state).
\end{align}
We can rewrite the classic Bellman eqution as
\begin{align}
	Q^\pi(\state, \action) = r(\state, \action) + \mathbb{E}_{\state', \epsilon}\left[Q^\pi(\state', f_\theta(\state', \epsilon))\right], \nonumber
\end{align}
and taking the gradients on both the sides yields
\begin{align}
	& \nabla_\theta Q^\pi(\state, \action) = \gamma \mathbb{E}_{\state', \epsilon}\left[	\nabla_{\action'}Q^\pi(\state', \action')\big|_{a' = f_\theta(\state', \epsilon)} \nabla_\theta f_\theta(\state', \epsilon) + \nabla_\theta Q^\pi(\state', \action') \big|_{a' = f_\theta(\state', \epsilon)} \right] \nonumber \\
	\implies&
	\bm{\Gamma}^\pi(\state, \action) = \gamma\mathbb{E}_{\state', \epsilon}\left[ \mathbf{g}_{REP}(\state', \action') + \bm{\Gamma}(\state', \action')\right],
\end{align}
where the \textsl{immediate reparametrization gradient} is
\begin{align}
	\mathbf{g}_{REP}(\state', \action')  = \nabla_{\action'}Q^\pi(\state', \action')\big|_{a' = f_\theta(\state', \epsilon)} \nabla_\theta f_\theta(\state', \epsilon).
\end{align}
\subsection{An Experiment with Continuous Action Space}

\begin{figure}
	\centering
	\includegraphics{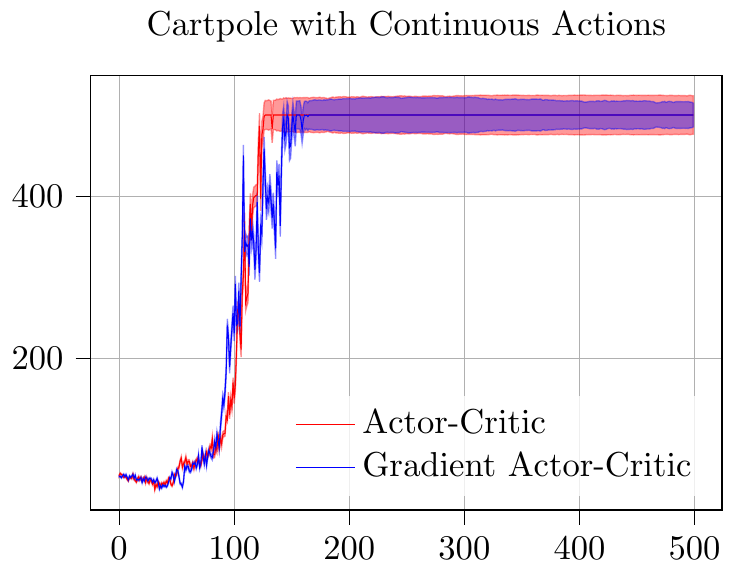}
	\caption{We run two algorithms: a classic actor-critic architecture and our gradient actor-critic architecture. On this task, both the algorithms exhibit similar performace, showing, nevertheless that our gradient actor-critic successfully solves the task.
		\label{fig:cartpole}}
\end{figure}

We demonstrate the applicability of the Gradient Actor-Critic method on a simple control task with continuous state-action spaces.
Our goal is two-fold, first to show that a simple heuristic allows for the use of neural network function approximation without an excessive computational cost, and the second to show that the proposed methodology can successfully solve a continuous control task.
We perform this demonstration using the continuous action Cartpole environment with a scalar action $u \in [-1, 1]$ which applies a lateral force on the cart for a one second period.
We cut off the episode after a maximum of 500 steps, then reinitialize the cart with a random velocity and the pole with a random pole angle and angular velocity.

To estimate the action-value function for the critic, we use a two hidden-layer neural network with tanh activations and 64 units per layer.
We feed the observable state and action into the neural network and have two heads attached to the penultimate layer, one head for the standard critic and another for the gradient critic.
The policy is likewise parameterized by a two hidden-layer neural network with tanh activations and 64 units per layer.
We swept the hyperparameters for both the Gradient Actor-Critic and Actor-Critic baseline, selecting the maximizing hyperparameter setting using 30 random seeds.
The swept hyperparameters are reported in the table below.
We then reran each algorithm for 100 random seeds for the maximizing hyperparameter setting in order to minimize maximization bias.

In Figure~\ref{fig:cartpole} we show that both the Gradient Actor-Critic and Actor-Critic methods are able to successfully learn a near-optimal policy on this task, with the optimal return being 500.
Although the Cartpole task is too simplistic to induce differences between these algorithms, it does highlight that the Gradient Actor-Critic method can easily solve a continuous control problem using neural network function approximation.
Both methods incurred near-identical computational cost on this problem setting, taking on average three minutes per run using a modern desktop processor.

\textbf{Hyperparameters for continuous control experiment:}
\begin{table}[h!]
	\begin{tabular}{|l|c|}
		\hline
		Optimizer & ADAM$(\beta_1 = 0.9, \beta_2=0.999)$ \\
		Target network moving average & $\{0.99, 0.9\}$ \\
		Learning rate for the critic & $\{0.1, 0.01, 0.001, 0.0001\}$ \\
		Learning rate for the actor & $\{0.1, 0.01, 0.001, 0.0001\}$ \\
		Eligibility Trace & $\{0.9, 0.75, 0.5, 0.1\}$ \\
		\hline
	\end{tabular}
\end{table}

\newpage
\section{Supplement to the Empirical Analysis}\label{app:empirical}
This section introduces some notes on the practical implementation of the algorithms, environments, and hyperparameters and settings used in the experiments. We finally present some complementary results to the one presented in the main paper.

\subsection{LSTD$\bm{\Gamma}$}\label{app:lstd-gamma}
Algorithm~\ref{alg:lstd-gamma} illustrates ``pure'' LSTD$\bm{\Gamma}$ ($\lambda=0$). This section discusses how to incorporate the eligibility traces in practice and how to write a simple `pytorch` spinnet to compute the gradient.

\textbf{Eligibility Trace.} To implement eligibility traces, we need a dataset where for each transition $s$ $a$ $r$ $s'$, we have also accompanied with $t$, a variable indicating the number of steps that occurred since the beginning of the current episode.
Hence, we first fit the matrix $\mathbf{A}_\pi$, and we compute the parameter matrix $\mathbf{G}$ and then, we compute the gradient as in Equation~\ref{eq:semi-traces}. A schematic representation of LSTD$\bm{\Gamma}$ can be found in Algorithm~\ref{alg:lambda-lstd-gamma}.

\begin{algorithm}[t]
	\caption{LSTD$\bm{\Gamma}$ \label{alg:lambda-lstd-gamma}}
	\begin{algorithmic}[1]
		\STATE \textbf{Input:} Set of features $\phi$, dataset $D$ of transitions $(\state_i, \action_i, r_i, \state_{i}', t_i )$ where $r_i$ are the rewards, $\state_i'$ the next states and $t_i$ is the time-step, policy $\pi_\theta$, learning rate $\eta$
		\STATE {$\hat{\mathbf{b}} =1/N \sum_i \phi(\state_i, \action_i) r_i$}
		\WHILE {not converged}
		\STATE Fore each $\state_i'$ sample $\action_i' \sim \pi_\theta(\cdot | \state_i')$
		\STATE {$\hat{\mathbf{A}} = 1/N \sum_i \phi(\state_i, \action_i)\left((\state_i, \action_i) - \gamma \phi(\state_i', \action_i')\right)^\intercal$}
		\STATE $\hat{Q}(\state, \action) = \phivec^\intercal(\state, \action)\hat{\omegavec}_{T\!D}$; $\hat{\omegavec}_{T\!D} = \hat{\mathbf{A}}^{-1} \hat{\mathbf{b}}$;
		\STATE {$\hat{\mathbf{B}} =1/N \sum_i \phi(\state_i, \action_i)\hat{Q}(\state_i', \action_i') \nabla_\theta\log\pi_\theta(\action_i'|\state_i')$}
		\STATE {$\hat{\bm{\Gamma}}(\state, \action) = \phivec^\intercal(\state, \action)\hat{\mathbf{G}}_{T\!D}$; $\hat{\mathbf{G}}_{T\!D} = \hat{\mathbf{A}}^{-1} \hat{\mathbf{B}}$}
		\STATE Sample $\state_i$ from dataset and $\action_i \sim \pi_\theta(\cdot|\state_i)$
		\STATE $\mathbf{g}_i = \hat{Q}(\state_i, \action_i) \nabla_\theta\log\pi_\theta(\action_i|\state_i)$
		\STATE $\theta \leftarrow\theta + \eta \lambda^{t_i} \gamma^{t_i} (\mathbf{g}_i + \hat{\bm{\Gamma}}(\state_i, \action_i))$
		\ENDWHILE
	\end{algorithmic}
\end{algorithm}

\textbf{Using Automatic Differentiation.} One can actually avoid to compute $\mathbf{G}$. When we look bach to Lemma~\ref{lemma:gradient-td}, we see that $\mathbf{G}_{T\!D} = \nabla_\theta \omegavec_{T\!D}$. Automatic differentiation via \texttt{pytorch} is actually able to derive that step automatically. Therefore, instead of computing $\mathbf{G}$ explicitly in the code, one can simply compute $\omegavec_{T\!D}$ and let the automatic differentiation tto find $\nabla_\theta \omegavec_{T\!D}$, as in Algorithm~\ref{alg:lambda-lstd-gamma-torch}.

\begin{algorithm}[t]
	\caption{LSTD$\bm{\Gamma}$ with Automatic Differentiation \label{alg:lambda-lstd-gamma-torch}}
	\begin{algorithmic}[1]
		\STATE \textbf{Input:} Set of features $\phi$, dataset $D$ of transitions $(\state_i, \action_i, r_i, \state_{i}' )$ where $r_i$ are the rewards and $\state_i'$ the next states, policy $\pi_\theta$, learning rate $\eta$
		\STATE {$\hat{\mathbf{b}} =1/N \sum_i \phi(\state_i, \action_i) r_i$}
		\WHILE {not converged}
		\STATE Fore each $\state_i'$ sample $\action_i' \sim \pi_\theta(\cdot | \state_i')$
		\STATE {$\hat{\mathbf{A}} = 1/N \sum_i \phi(\state_i, \action_i)\left((\state_i, \action_i) - \gamma \phi(\state_i', \action_i')\right)^\intercal$}
		\STATE $\hat{Q}(\state, \action) = \phivec^\intercal(\state, \action)\hat{\omegavec}_{T\!D}$; $\hat{\omegavec}_{T\!D} = \hat{\mathbf{A}}^{-1} \hat{\mathbf{b}}$;
		\STATE Sample $\state_t$ from dataset and $\action_i \sim \pi_\theta(\cdot|\state_i)$
		\STATE $\mathbf{g}_i = \hat{Q}(\state_i, \action_i) \nabla_\theta\log\pi_\theta(\action_i|\state_i)$
		\STATE $\theta \leftarrow\theta + \eta\lambda^{t_i} \gamma^{t_i} \nabla_\theta \hat{Q}(\state_i, \action_i)$
		\ENDWHILE
	\end{algorithmic}
\end{algorithm}
\subsection{TDRC$\bm{\Gamma}$.}\label{app:tdrc-gamma} This algorithm, described in Algorithm~\ref{alg:tdrc-gamma}, uses TDRC to estimate both critic and gradient critic.
To do so, we simply replace the semi-gradient TD update rule with the following TDRC update
\begin{align}
	& \delta_t = \Reward_t + \gamma \hat{Q}^\pi_{t}(\State_{t+1}, \Action_{t+1}) - \hat{Q}^\pi_{t}(\State_{t}, \Action_{t}) \nonumber \\
	& \bm\chi_{t+1} = \bm\chi_t + \alpha_t \phivec_t \left(\delta_t - \phivec_t^\intercal \bm\chi_t\right) - \alpha_t\beta \bm\chi_t \nonumber \\
	& \omegavec_t  = \omegavec_t + \alpha_t \phivec \delta_t - \alpha \gamma \phivec_t' \phivec^\intercal_t \bm\chi_t, \label{eq:tdrc}
\end{align}
where $\chi_t$ are a secondary set of weights to perform gradient correction and $\beta$ is the TDRC regularization factor.

Similarly, we can estimate the gradient critic with a vector form of TDRC,
\begin{align}
	& \bm{\varepsilon}_{t} =  \gamma \hat{Q}^\pi_{t}(\State_{t+1}, \Action_{t+1}) \nabla_\theta \log \pi_\theta(\Action_{t+1} | \State_{t+1})\nonumber\\
	& \qquad  +  \gamma \hat{\bm\Gamma}^\pi_{t}(\State_{t+1}, \Action_{t+1}) - \hat{\bm\Gamma}^\pi_{t}(\State_t, \Action_t) \nonumber \\ 
	& \mathbf{H}_{t+1} =  \mathbf{H}_{t} + \alpha_t \phivec\left(\bm\varepsilon_t^\intercal  - \phivec^\intercal_t \mathbf{H}_t\right) - \alpha_t \beta \mathbf{H}_t \nonumber \\
	& \mathbf{G}_{t+1} = \mathbf{G}_t + \alpha_t \bm\phi\bm\varepsilon_t^\intercal -\alpha\gamma\bm\phi'\bm\phi^\intercal \mathbf{H}_t, \label{eq:tdrc-gamma}
\end{align}
where $\hat{\bm\Gamma}_t^\pi(S, A) = \phivec^\intercal(S, A)\mathbf{G}_t$ and $\hat{Q}$ is an estimate of the critic. $\mathbf{H}_t$ have the same role as $\bm{\chi}_t$ in \eqref{eq:tdrc}. The samples $S_t, A_t, S_{t+1}, A_{t+1}$ are sampled i.i.d.\ according to $\zeta$.

Because the critic and gradient critic estimations have no circular dependencies, we can easily prove convergence of the gradient critic to $\hat{\bm{\Gamma}}_{T\!D}^\pi$ by simply allowing TRDC to first converge to $\hat{Q}_{T\!D}^\pi$ and subsequently iterating \eqref{eq:tdrc-gamma} using $\hat{Q}_{T\!D}^\pi$, converging therefore to $\hat{\bm{\Gamma}}_{T\!D}^\pi$.
However, such an approach is not practical. To obtain faster convergence, we propose to interleave both the updates in \eqref{eq:tdrc}, \eqref{eq:tdrc-gamma}, and of the target polcy. We call this algorithm TDRC$\bm{\Gamma}$ (Algorithm~\ref{alg:tdrc-gamma}).

To have as few hyperparameters as possible, we set the same learning rate for both the critic and the gradient critic. Across all the experiments, we use $\beta=1$.
To be precise, in \texttt{Imani's MDP}, one could avoid using a full-gradient TD technique (like TDC, GTC, ...) since the critic features are perfect. However, we preferred to maintain consistency between different experiments.

\subsection{Imani's MDP.}\label{app:limani-mdp} There are a few choices that can be made to implement this MDP. We opted to implement this MDP as a four-state MDP where the terminal state is absorbing. We did this because our current code computes the policy gradient in closed form without knowing terminal states. This modification is not an issue. Making $T$ an absorbing state changes the discounted stationary distribution, leaving the ratio of visitation between $S_0$, $S_1$, and $S_2$ unchanged, which is, after all, what matters. Furthermore, the gradient on the absorbing state is always $0$.

To allow generality, our policy, therefore, accepts the input of $4$ different states, and, since the possible actions per state are two, the tabular policy is encoded with $8$ parameters. In the presence of state-aliasing, however, when the MDP is in state $S_2$, state $S_1$ is fed to the policy instead. For this reason, from the policy perspective, state $S_2$ is never visited, causing the gradient of the parameters that correspond to state $S_2$ to be always zero.

These implementation choices do not change the math and the effects of the original MDP of Imani's et al.

The parameters that matter are $\theta_0, \theta_1, \theta_2$ and $\theta_3$, corresponding to state $S_0$ and $S_1$ (which is aliased with $S_2$).

\subsection{Bias-Variance Tradeoff in Figure~\ref{fig:imani-bias}b, c, and d}\label{app:bias-variance}
\texttt{Imani's MDP} has a closed-form solution of the policy gradient. We use this solution to compute the bias of the estimators. While the experiment's setting has been already described in the paper, here we provide fewer details on how the bias and the variance have been estimated.

We build both variance and bias estimates for each value of $\lambda$ by sampling $20$ instances of the estimators (e.g., running $20$ times the algorithm to estimate the gradient). After, we compute the squared bias and the variance per component, i.e.,
\begin{align}
	\hat{\mathbf{b}} = \left(\frac{1}{20}\sum_{i=1}^{20} \left(\hat{\mathbf{g}}_i - \nabla_\theta J(\theta)\right)\right)^2; \quad \hat{\mathbf{v}} = \frac{1}{20}\sum_{i=1}^{20} \left(\hat{\mathbf{g}}_i - \overline{\mathbf{g}} \right)^2, \nonumber
\end{align}
where $\mathbf{g}_i$ are the single estimates of the gradient, $\nabla_\theta J(\theta)$ the true gradient, and $\overline{\mathbf{g}}$ the empirical average of the gradient estimate.
The vectors of empirical bias $\mathbf{b}$ and variance $\mathbf{v}$ are then transformed to scalars by taking the mean over the components, i.e.,
\begin{align}
	\hat{b} = \frac{1}{8}\sum_{i}^8 \hat{\mathbf{g}}_i; \quad \hat{v} = \frac{1}{8}\sum_{i}^8 \hat{\mathbf{v}}_i. \nonumber
\end{align}
Now, $\hat{b}$ and $\hat{v}$ are also estimates. Therefore, we repeat this process $50$ times to compute an empirical average of the estimates and build confidence intervals. Therefore, for each value of $\lambda$, we compute $1000$ estimates of the gradient. We show the estimate both on a circular plot, which shows the ground truth and the single estimates compactly, and we also report the single estimates of $\{\partial/\partial \theta_i\}_{i=0}^3$ (since the remaining partial derivatives are all equal to zero).

\begin{figure}[H]
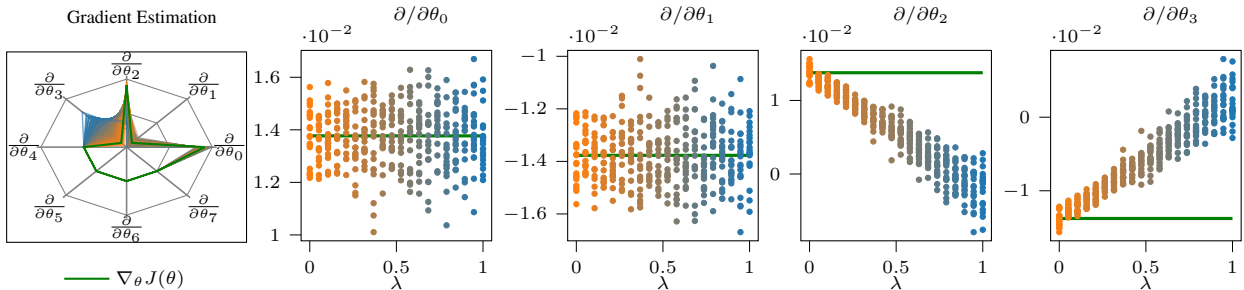

	\input{figures/gradient_lstd_gamma.tex}
	\hspace{-0.3cm}
	\input{figures/gradient-dim-0.tex}
	\hspace{-0.3cm}
	\input{figures/gradient-dim-1.tex}
	\hspace{-0.3cm}
	\input{figures/gradient-dim-2.tex}
	\hspace{-0.3cm}
	\input{figures/gradient-dim-3.tex}
	\hspace{-0.3cm}
	\caption{On the left, a plot with all the gradient estimates. In orange color, we have low values of $\lambda$ (hence, full use of the gradient critic); in blue color, we have the semi-gradient estimator. We denote in green the ground truth. In the plots on the left, we show the actual gradient estimates for the first four parameters. Parameters $\theta_2$ and $\theta_3$ are critical, as they are subject to state aliasing. The gradient critic delivers an unbiased estimate, while semi-gradient exhibits high bias.}
\end{figure}

\subsection{LSTD$\bm{\Gamma}$ - Figure~\ref{fig:learning-curves}a}
The performance of LSTD$\bm{\Gamma}$ on \texttt{Imani's MDP} has been shown on Figures~\ref{fig:learning-curves}a. In this experiments, we sampled a dataset of $500$ using the behavioral policy, and we applied LSTD$\bm{\Gamma}$ for $1000$ steps. More in particular, the estimated gradient has been used with Adam (with learning rage $0.01$). At each step, the return of the target policy is computed in closed form.
We inspect $20$ values of $\lambda$ in the interval $[0, 1]$, performing $10$ different indipendent runs of the algorithm to appreciate confidence intervals at $95\%$. Since most values of $\lambda$ tend to have similar return, we defided both to show the final performance (at the $1000$th iteration), and a few learning curves.
We also computed the learning curve of pure semi-gradient and pure LSTD$\bm{\Lambda}$ using the gradients in closed-form. Figure~\ref{fig:closed-form} depicts the learning curve obtained for a fewer values of $\lambda$.


\subsection{TDRC$\bm{\Gamma}$ - Figure~\ref{fig:learning-curves}c}
\label{app:gamma-rc-imani}
\begin{figure}
	\centering	\includegraphics{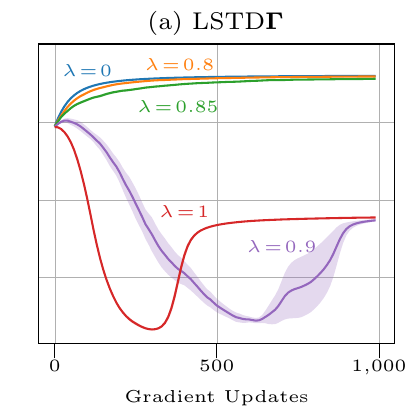}
	\input{figures/rc-last-imani.tex}
	\caption{(a) Learningt curves of LSTD$\bm\Gamma$ on Imani's MDP (b) TDRC$\bm\Gamma$ applied on \texttt{Imani's MDP} and evaluated at the last training step. The hyper-parameter $\lambda$ behaves similarly to LSTD$\bm\Gamma$ in Figure~\ref{fig:learning-curves}a \label{fig:lambda-rc-gamma-imani}\label{fig:closed-form}.}
\end{figure}

This figure has been produced by running TDRC with parameters $\beta=1$, $\alpha=0.1$, and Adam with a learning rate $0.001$ for the actor update. Surprisingly, the curve is almost identical to the one obtained in Figure~\ref{fig:learning-curves}b. The confidence intervals have been obtained by running $20$ instances for each value of $\lambda$. Moreover, in Figure~\ref{fig:lambda-rc-gamma-imani}, we show the performance at the last iteration step for $20$ values of $\lambda$ in the range $[0, 1]$. Interestingly, $\lambda$ behaves similarly to LSTD, as in Figure~\ref{fig:learning-curves}a.

\textbf{OffPAC:}

\begin{tabular}{|l|c|}
	\hline
	Learning rate for the critic & $0.1$ \\
	Learning rate for the actor & $0.001$ \\
	GDT Regulariztion & $0.1$ \\
	Eligibility Trace & $0.1$ \\
	\hline
\end{tabular} 

\textbf{ACE($\eta=1$):}

\begin{tabular}{|l|c|}
	\hline
	Learning rate for the critic & $0.1$ \\
	Learning rate for the actor & $0.001$ \\
	Entropic Regularization & $0$ \\
	GDT Regulariztion & $0.1$\\
	Eligibility Trace & $0.1$ \\
	\hline
\end{tabular} 

Note that we estimate the critic using GDT both for OffPAC and ACE.

\textbf{TDRC$\bm\Gamma$:}

\begin{tabular}{|l|c|}
	\hline
	Learning rate for the value critic & $0.1$ \\
	Learning rate for the actor critic & $0.1$ \\
	Learning rate for the actor & $0.001$ \\
	TDRC Regulariztion & $1.0$\\
	\hline
\end{tabular} 

\subsection{Experiments on Random MDPs- Figure~\ref{fig:random-mdps}c and d}
\label{app:tdrc-random-mdp}
To enerate the transition and the reward model, we first sample a uniform vector, and then we feed it in a soft-max function
\begin{align}
	\mathrm{SoftMax}(\mathbf{x})_i = \frac{\exp{T x_i}}{\sum_j\exp{T x_j}},
\end{align}
where the temperature $T$ controls the entropy of the overall distribution. With high $T$ we tend to have sparse reward and deterministic transition, while with low $T$, uniform transitions and reward model. In our experiments, where we use $30$ states and $2$ action, a temperature $T=10$ seems to be a good balance to generate interesting models. As explained in the main paper, when interacting with the MDP, the agent observes the reward with an addition of a Gaussian noise with standard deviation of $0.1$. 

In the following, we describe the setting used for this experiment.

\textbf{OffPAC:}

\begin{tabular}{|l|c|}
	\hline
	Learning rate for the critic & $(|\Sset||\Aset|)^{-1}$ \\
	Learning rate for the actor & $10^{-2} (|\Sset||\Aset|)^{-1}$ \\
	GDT Regulariztion & $0.1$ \\
	Eligibility Trace & $0$ \\
	\hline
\end{tabular} 

\textbf{ACE($\eta=1$):}

\begin{tabular}{|l|c|}
	\hline
	Learning rate for the critic & $(|\Sset||\Aset|)^{-1}$ \\
	Learning rate for the actor & $10^{-2}(|\Sset||\Aset|)^{-1}$ \\
	Entropic Regularization & $0$ \\
	GDT Regulariztion & $0.1$\\
	Eligibility Trace & $0$ \\
	\hline
\end{tabular} 

Note that we estimate the critic using GDT both for OffPAC and ACE.

\textbf{TDRC$\bm\Gamma$:}

\begin{tabular}{|l|c|}
	\hline
	Learning rate for the value critic & $(|\Sset||\Aset|)^{-1}$ \\
	Learning rate for the actor critic & $(|\Sset||\Aset|)^{-1}$ \\
	Learning rate for the actor & $10^{-2} (|\Sset||\Aset|)^{-1}$ \\
	TDRC Regulariztion & $1.0$\\
	\hline
\end{tabular} 

\end{document}